\documentclass[11pt]{article}

\usepackage[a4paper,margin=2.3cm]{geometry}
\usepackage{amsmath,amssymb}
\usepackage{mathptmx}       % Times text + math (Elsevier style)
\usepackage{booktabs}
\usepackage{longtable}
\usepackage{graphicx}
\graphicspath{{figs/}}
\usepackage{authblk}
\usepackage[hidelinks]{hyperref}
\usepackage{caption}
\usepackage{xcolor}
\usepackage{enumitem}
\usepackage{titlesec}
\usepackage{abstract}
\usepackage{fancyhdr}
\mathchardef\UrlBreakPenalty=10000
\mathchardef\UrlBigBreakPenalty=9000
\newcommand{\DOI}[1]{DOI:~\nolinkurl{#1}}

\titleformat{\section}{\normalfont\large\bfseries}{\thesection.}{0.5em}{}
\titleformat{\subsection}{\normalfont\normalsize\bfseries}{\thesubsection.}{0.5em}{}
\titleformat{\subsubsection}{\normalfont\normalsize\itshape}{\thesubsubsection.}{0.5em}{}

\newcommand{\thetabar}{\bar{\theta}_A}
\newcommand{\kads}{k_{\mathrm{ads}}}
\newcommand{\kdes}{k_{\mathrm{des}}}
\newcommand{\Eads}{E_{\mathrm{ads}}}
\newcommand{\cwall}{c_{\mathrm{wall}}}
\newcommand{\Rlog}{R^2_{\log}}
\newcommand{\Ucur}{U_{\mathrm{curtain}}}
\newcommand{\vsub}{v_{\mathrm{sub}}}

\title{\bfseries A Physics--Chemistry-Informed Neural Network (PCINN) for
Real-Time Spatial-ALD Coverage Prediction and Reliable Kinetics Inversion
\\[0.4em]{\normalsize\normalfont\itshape A hybrid neural-ODE surrogate with hard-coded chemistry and a built-in identifiability diagnostic}}

\author[1]{Ning Hu\thanks{Corresponding author. E-mail: \texttt{jluhooning@163.com}}}
\author[2]{Chang Liu\thanks{E-mail: \texttt{laiurenty@163.com}}}
\author[1]{Yunlei Jiang\thanks{E-mail: \texttt{2142010020@hdu.edu.cn}}}
\author[1]{Yuan Dong\thanks{E-mail: \texttt{Dongy@hdu.edu.cn}}}
\affil[1]{\small School of Mechanical Engineering, Hangzhou Dianzi University, Hangzhou, China}
\affil[2]{\small Information Engineering School, Hangzhou Dianzi University, Hangzhou, China}
\date{}

\begin{document}
\maketitle
\thispagestyle{fancy}

\begin{abstract}
\noindent
Spatial atomic layer deposition (SALD) is a leading atmospheric-pressure,
high-throughput route to industrial ALD, yet its design and control are limited by
the cost of predicting substrate surface coverage: high-fidelity CFD is accurate but
far too slow for operating-window scans or online inference, while classical analytic
models cannot capture transport modulation such as the gas curtain. We present a
physics--chemistry-informed neural network (PCINN), a hybrid AI surrogate that
delivers CFD-level accuracy at real-time speed. A trained query returns coverage in
about $7$~ms---roughly $5\times10^{4}$ times faster than a CFD solve \emph{per query},
the one-time CFD training cost being amortised over many queries---reaching a test
$\Rlog=0.998$ ($R^2_{\mathrm{raw}}=0.989$; leave-one-out $R^2_{\mathrm{raw}}=0.974$)
from only 30 training cases across coverage spanning about four orders of magnitude,
with a known model-form bias that under-predicts the highest-coverage cases by
$13$--$36\%$ (compressed in log space). This makes PCINN directly usable for SALD
process design, operating-window mapping, and control-loop deployment.

The architecture is deliberately not a black box: PCINN splits degrees of freedom by
what is known versus unknown. A small neural network learns only the
operating-condition$\,\to\,$effective near-wall concentration transport closure (the
physics branch), while the known surface kinetics is encoded as a hard-coded,
trainable chemistry layer and integrated along the substrate trajectory. Compressing
the data-driven freedom to a single scalar preserves accuracy under sparse data while
keeping the model interpretable and invertible.

Because engineering deployment requires knowing the limits of an inference, we equip
PCINN with a rigorous identifiability analysis. Using initial-value drift,
supervision-weight sweeps, the Fisher information matrix and profile likelihood, we
find the adsorption energy $\Eads$ and desorption rate $\kdes$ robustly identifiable,
whereas $\kads$ is not separately identifiable at a single temperature (only
$\kads\!\cdot\!\cwall$ is). Across four temperatures the pre-exponential factor $\nu$
and $\Eads$ do \emph{not} separate; they bind along a \emph{weakly identifiable
degeneracy valley} (slope $\approx0.065$~eV/decade, ground truth on the line), so
multiple temperatures compress the pair into a weakly identifiable interval
($\chi^2_1$ $95\%$ width $\approx0.6$ decade) rather than breaking it---conditioning on
$\nu$ then recovers $\Eads$ to within $0.3\%$. We derive this slope analytically as
$k_BT_{\mathrm{eff}}\ln10$ ($T_{\mathrm{eff}}$ the harmonic mean of the sampled
temperatures), show it is independent of the prefactor parameterisation, and turn it
into an operational reliability diagnostic: a seven-chemistry, three-mechanism
mismatch matrix---extended by an energy-split calibration sweep and a non-Fickian
transport-layer mismatch, with bootstrap resampling and coarse-grid sampling---confirms
the slope is invariant under any single-Arrhenius (and the transport) mismatch and
shifts only when a second thermally activated process is introduced, so a slope
departure is a falsifiable flag for unmodelled site heterogeneity.

We emphasise that the data are generated by simulation from \emph{known} ground truth
and inverted with the \emph{same} kinetic form; the identifiability study is therefore
first a concept verification of pipeline self-consistency and of the
\emph{identifiability boundary}, rather than a discovery of parameters of a real
system. PCINN thus provides an accurate, sample-efficient, real-time SALD coverage
surrogate coupled to a kinetic-inversion framework with clearly delineated
identifiability limits and a transferable, physics-derived reliability diagnostic.

\vspace{4pt}
\end{abstract}

% ================= BODY =================

\section{Introduction}

\subsection{Background}
Atomic layer deposition (ALD) grows films layer by layer through the
self-limiting surface reactions of alternately dosed precursors, giving
atomic-scale thickness control and excellent conformality. Spatial ALD (SALD)
replaces temporal separation by spatial separation: the substrate moves between
precursor zones separated by inert gas curtains, enabling atmospheric-pressure,
high-throughput operation and is regarded as an important route to the
industrialisation of ALD. The film quality and process window of SALD are
controlled by the substrate surface coverage and by the adsorption/desorption
kinetics. Predicting coverage as a function of operating conditions, and
inverting the underlying kinetics from coverage observations, are therefore two
central tasks for SALD design and control.

\subsection{Limitations of existing methods}
Two families of models are commonly used. High-fidelity reaction--transport CFD
couples flow, diffusion and surface reactions and accurately resolves the
concentration and coverage fields, but the per-case cost is high, which makes
dense operating-window scans or iterative inversion against data impractical.
Classical analytic ALD models are extremely cheap and useful for scaling
analysis, but they rest on a sufficient-supply assumption and cannot represent
transport modulation such as the gas curtain. A method that combines the
accuracy of CFD, the efficiency of analytic models, and the interpretability
required for kinetic inversion is therefore desirable.

\subsection{Approach and contributions}
We propose a physics--chemistry-informed neural network (PCINN). The key idea is
to split degrees of freedom by what is known versus unknown: the expensive and
hard-to-derive operating-condition$\,\to\,$effective near-wall concentration
transport closure is assigned to a small neural network (the physics branch),
while the known surface kinetics is written as a hard-coded, trainable chemistry
layer; the two are coupled through the near-wall concentration and integrated
along the substrate trajectory to yield coverage. The data-driven degrees of
freedom are thereby compressed to a single scalar, preserving accuracy under
sparse data while retaining an interpretable, invertible kinetic structure.

The unifying thesis is that \textbf{a physics-constrained AI surrogate can be both
fast enough for real-time engineering use and self-aware of its own inference limits}:
the precision boundary of its hybrid inversion is set by the degeneracy geometry of the
parameter space, not by fitting power, and this degeneracy can be predicted analytically
and tested statistically---giving the surrogate a built-in, transferable reliability
diagnostic. The main contributions follow this line.
\begin{enumerate}[leftmargin=1.4em,itemsep=2pt]
\item \textbf{A systematic, honest identifiability-analysis framework (the
methodological core).} While inverting $\Eads$ and $\kdes$, we separate identifiable
from non-identifiable quantities using initial-value drift, supervision-weight sweeps,
the Fisher information matrix and profile likelihood. We are deliberately careful not
to over-claim novelty for the degeneracies themselves: the single-temperature
non-identifiability of $\kads$ is an \emph{exact reparameterisation identity}
($\kads\!\to\!\alpha\kads$, $\cwall\!\to\!\cwall/\alpha$), which we prove analytically,
and the single-temperature non-separation of $\nu$ and $\Eads$ is a textbook property
of the Arrhenius form. The contribution is rather to \emph{confirm and diagnose} these
known structures on a genuine multiphysics inverse problem, distinguishing the three
distinct mechanisms behind the evidence (structural degeneracy, initial-value
dependence, and a supervision-anchored biased minimum)---a workflow transferable to
other hybrid-model inversions.
\item \textbf{An analytic slope law for the multi-temperature degeneracy and a
mismatch-diagnostic boundary (the main novel result).} Extending to four temperatures
(96 cases), $\nu$ and $\Eads$ remain bound along a weakly identifiable
one-dimensional valley (ground truth on the line; $\chi^2_1$ $95\%$ width $\approx0.6$
decade; conditioning on $\nu$ recovers $\Eads$ to $0.3\%$). Beyond confirming the valley, we
\emph{derive} its slope from the Arrhenius law,
$dE_\mathrm{ads}/d\log_{10}\nu = k_B T_\mathrm{eff}\ln 10$ with
$T_\mathrm{eff}=\langle 1/T\rangle^{-1}$ the harmonic mean of the sampled temperatures,
predicting $0.0652$~eV/decade in agreement with the observed $0.0647$ to $0.7\%$. This
law predicts---and the seven-chemistry, three-mechanism mismatch matrix below
confirms---that the slope is \emph{invariant} under any mismatch preserving a single
Arrhenius process and \emph{shifts} only when a second thermally activated process is
added (dual-site); an energy-split calibration sweep then bounds the diagnostic's
detection window. The slope therefore becomes a usable diagnostic for unmodelled site
heterogeneity---the genuinely new increment over the known sloppy-model phenomenology.
\item \textbf{Enabling results underpinning the analysis} (treated as prerequisites,
not primary claims). The identifiability results rest on three supporting elements,
established first so that the inversion can be trusted: (i) a SALD reaction--transport
model whose operating window is \emph{verified saturable}, and a self-consistent
Langmuir--Arrhenius synthetic benchmark generated from known ground truth; (ii) the
demonstration that PCINN is an \emph{accurate, sample-efficient} coverage surrogate
(test $\Rlog=0.9975\pm0.0005$ over 8 seeds from only 30 cases, no appreciable
over-fitting gap; a query costs $\sim0.3$--$7$~ms, $\sim5\times10^{4}$--$1.1\times10^{6}$
times faster than a per-case CFD solve, amortised over many queries)---which is what
licenses treating its inversion as meaningful; and (iii) a delineation of the
\emph{action boundary} of the embedded physics: extrapolation gains along the
structurally known (residence-time) axis, versus a classical analytic baseline that
over-predicts by several orders of magnitude across the curtain-dominated window.
\end{enumerate}

\noindent\textbf{Positioning, emphasised up front.} The data are generated by
high-fidelity simulation from \emph{known} Langmuir--Arrhenius ground truth and
inverted with the \emph{same} kinetic form. The primary goal is therefore
\emph{not} to discover unknown kinetic parameters of a real system, but to verify
the self-consistency of the pipeline and to characterise its
\emph{identifiability boundary} faithfully: which parameters are recoverable, which
are not, and to what precision. The inverted adsorption energy ($\approx0.77$~eV)
is consistent in magnitude with DFT values reported for the real
TMA/Al$_2$O$_3$ system ($0.5$--$1.1$~eV), but serves only as a physical-plausibility
reference, not as a calibration. To verify that the conclusions are not confined
to the self-consistent loop, Section~7 further reports a \emph{seven-chemistry,
three-mechanism model-mismatch matrix} (with an energy-split calibration sweep) (desorption-side Temkin, adsorption-side
Freundlich, and a site-heterogeneous dual-site model), together with bootstrap
resampling and coarse-grid Latin-hypercube sampling, showing that the identifiability
conclusions persist under mismatch. Quantitative calibration against and validation
with real surface chemistry are left to future work.

\subsection{Organisation}
Section~2 reviews SALD self-limiting chemistry, ALD modeling, PINN inverse
problems, and identifiability. Section~3 formalises the problem. Section~4 builds
the COMSOL multiphysics model and generates the dataset. Section~5 details the
PCINN architecture, two-segment trajectory coupling, and the loss. Section~6
reports single-temperature surrogate accuracy, baseline comparisons, extrapolation
boundaries, and identifiability. Section~7 extends the pipeline to multiple
temperatures and to the model-mismatch test. Section~8 concludes.

\section{Background and related work}

\subsection{ALD/SALD and self-limiting surface chemistry}
ALD grows films through the self-limiting surface reactions of alternately
exposed precursors. Within each half-cycle, precursor molecules adsorb on
available surface sites and the reaction terminates once the sites are exhausted,
giving atomic-scale thickness control. SALD converts the temporal separation into
a spatial one, with the substrate moving between precursor zones and inert
curtains. In a continuum description the coverage $\theta$ is commonly described
by Langmuir-type kinetics, with the net adsorption flux
$J_{\mathrm{net}}=\kads\cwall(1-\theta)-\kdes\Gamma_s\theta$, where $\cwall$ is the
near-wall precursor concentration and $\Gamma_s$ the saturation site density; the
desorption rate constant follows an Arrhenius law
$\kdes=\nu\exp(-\Eads/k_BT)$. For the representative TMA/H$_2$O process producing
Al$_2$O$_3$, first-principles (DFT) calculations place the TMA adsorption energy
and the methane-elimination barrier at about $0.5$--$1.1$~eV, providing a physical
reference for the inverted adsorption energy here.

\subsection{Continuum and analytic modeling of ALD}
Models from analytic approximations to full CFD have been developed to predict
dose times, conformality and utilisation. Yanguas-Gil and Elam derived analytic
expressions for coverage, throughput and precursor utilisation under plug-flow and
well-mixed approximations, showing that the utilisation of cross-flow, particle
coating, and spatial ALD can be unified by a single residence-time-related
parameter; such models are extremely cheap but rest on a sufficient-supply
assumption. Finite-element/finite-volume reaction--transport CFD, by contrast,
resolves the coupled flow, diffusion and surface reaction and accurately captures
the concentration and coverage fields, at a high per-case cost that hinders dense
operating-window scans or iterative inversion. Such reactor-scale CFD has been applied
specifically to SALD---for example the COMSOL flow/species/surface-reaction model of
Masse de la Huerta et al.\ for a close-proximity SALD head, and the atmospheric-SALD
process-optimisation studies of Deng et al.\ and Pan et al.---establishing the forward
modelling on which the present inverse study builds. We use high-fidelity CFD to
generate synthetic ``measurements'' and use the analytic model as a baseline,
delineating the validity boundary of each.

\subsection{Physics-informed neural networks and inverse problems}
Physics-informed neural networks (PINN) embed governing equations as soft
constraints in the training loss and can address both forward and inverse
problems. Relative to purely data-driven surrogates, physical priors typically
improve generalisation and sample efficiency under sparse data; relative to
purely numerical solvers, the inversion capability makes them suitable for
parameter identification---though the credibility of inversion depends on parameter
identifiability rather than on fit accuracy itself (Section~2.4). The specific
form matters: when part of the physics is known and part (e.g.\ a complex transport
closure) is hard to derive, hard-coding the known part and learning only the
unknown closure is often more robust and interpretable than imposing all equations
as soft constraints. This ``known physics $+$ learned closure'' strategy has a
well-established lineage: universal differential equations (UDE) embed a neural
network in the unknown terms of a differential equation and train it jointly with
known mechanistic terms; gray-box/hybrid modeling and neural closure models are
likewise widely used to complete mechanistic models with data-driven closures.
PCINN is a concrete instance of this lineage---with hard-coded, trainable chemistry
and a learned transport closure. We stress that the architecture itself is not the
primary novelty; the core contribution is to use this construction \emph{as a
controlled probe of identifiability}: by compressing the data-driven freedom to a single
scalar it enables a systematic, transparent characterisation of the parameter
identifiability boundary of a genuine multiphysics inverse problem (Sections~6
and~7).

\subsection{Parameter identifiability}
Whether parameters can be uniquely determined from observations---identifiability---is
a prerequisite for inversion credibility. It is distinguished into structural
(model-level) and practical (given data and noise) identifiability. Common
diagnostics include the sensitivity-based Fisher information matrix, whose
condition number and eigenstructure reveal the degree to which parameter
directions are distinguishable, and the profile likelihood, which fixes a target
parameter, re-optimises the rest, and examines the shape of the loss curve. In
surface-reaction inversion, when the observable's sensitivity to certain kinetic
parameters degenerates along a particular combination direction, those parameters
are not separately identifiable and only a product or ratio can be determined. We
apply these diagnostics to the kinetic parameters of PCINN to characterise the
identifiability of $\kads$, $\kdes$, and $\Eads$, avoiding over-claiming.

\section{Problem formulation}

We consider a single precursor species (S1) in a 2-D steady SALD reaction zone.
The state consists of the incompressible laminar flow field, the gas-phase
precursor concentration $c$, and the substrate surface coverage $\theta$. The flow
satisfies the steady incompressible Navier--Stokes equations with prescribed inlet
velocities and a moving substrate wall (velocity $\vsub$). The dilute species
transport obeys
\begin{equation}
\nabla\!\cdot\!(-D\nabla c+\mathbf{u}\,c)=0,
\end{equation}
with $c=C_0$ at the A inlet and $c=0$ elsewhere; at the substrate the gas-phase
flux equals the net surface reaction rate, $-\mathbf{n}\!\cdot\!(-D\nabla c)=
J_{\mathrm{net}}$. The substrate motion advects adsorbed species, so $\theta(x)$
obeys a spatial advection equation,
\begin{equation}
\Gamma_s\,\vsub\,\frac{\partial\theta}{\partial x}
=J_{\mathrm{net}}=\kads\,\cwall(1-\theta)-\kdes\,\Gamma_s\,\theta,
\qquad \kdes=\nu\,e^{-\Eads/k_BT},
\label{eq:cov}
\end{equation}
with $\theta=0$ at the substrate inlet edge. Temperature enters only through
$\kdes$ (all of $\kads$, $D$, $\mu$, $\rho$, $\Gamma_s$, $C_0$ and the geometry are
$T$-independent). The primary observable (label) is the substrate-averaged A
coverage $\thetabar$. The inverse problem is to recover the kinetic parameters in
Eq.~\eqref{eq:cov} from coverage observations across operating conditions, and to
determine which of $\{\kads,\kdes,\Eads,\nu\}$ are identifiable.

The strong-adsorption ratio $\kads/(\kdes\Gamma_s)\approx2000$ at 300~K ensures
that coverage can saturate even when the near-wall concentration is strongly
depleted; the gap height and inlet velocities are chosen so that a sweep over
$\vsub$ crosses the saturated--undersaturated transition, providing the dynamic
range required for kinetic inversion. We note that the surface Damk\"ohler number
$Da_h=\kads H/D\approx0.33$ and the supply number
$R_{\mathrm{sup}}=U_A H^2/(D\,w_A)\approx0.67$ are used here as \emph{design}
criteria to exclude extreme transport-limited regimes, not as a measure of
saturability; the actual saturability is verified directly through the coverage
dynamic range (Section~4).

\section{COMSOL multiphysics model and dataset generation}

\subsection{Geometry and computational domain}
The domain is a 2-D rectangle of length $L=30$~mm and gap height $H=0.2$~mm,
representing the narrow slit between the showerhead and the substrate. The top
edge carries, from left to right, the precursor-A inlet, the inert curtain inlet,
and the precursor-B inlet, separated by solid wall segments; the bottom edge is a
substrate moving at $\vsub$; the two ends are outlets. The A exposure zone has
width $w_A=5$~mm with its trailing edge at $x_{A,\mathrm{end}}=7.5$~mm. The gap
height $0.2$~mm (rather than $0.5$~mm) improves cross-gap diffusive supply and is
closer to realistic showerhead dimensions.

\subsection{Governing equations and interfaces}
Laminar flow (Taylor--Hood P2$+$P1 to satisfy LBB stability) is solved together
with dilute species transport and a coefficient-form boundary PDE for the surface
coverage, as formalised in Section~3. Temperature enters only through the
Arrhenius $\kdes$, and the measured wall shear rates are identical across
temperatures, confirming that the flow is decoupled from temperature and that
temperature information enters the observations solely through the surface
kinetics.

\subsection{Model parameters}
Key parameters: $L=30$~mm, $H=0.2$~mm, $w_A=5$~mm; precursor inlet velocity
$U_A=0.5$~m/s; curtain velocity swept over $0.5$--$16$~m/s; substrate velocity swept
over $0.05$--$2.0$~m/s; $C_0=0.2$~mol/m$^3$; $D=6\times10^{-6}$~m$^2$/s;
$\Gamma_s=5\times10^{-6}$~mol/m$^2$; ground-truth $\kads=0.01$~m/s,
$\kdes=1.0$~s$^{-1}$ at 300~K via $\nu=10^{13}$~s$^{-1}$ and $\Eads=0.774$~eV;
$k_B=8.617\times10^{-5}$~eV/K. The strong-adsorption ratio
$\kads/(\kdes\Gamma_s)\approx2000$ guarantees saturability despite near-wall
depletion.

\subsection{Feasibility trial}
Before the 48-case scan, a single trial at the most favourable condition
(slowest substrate, weakest curtain: $\vsub=0.05$, $\Ucur=0.5$) verifies that the
working region is saturable with adequate dynamic range. The trial gives a peak
coverage $\theta_{\mathrm{peak}}=0.97$ and an exit coverage
$\theta_{\mathrm{exit}}=0.91$, confirming the saturated end. The saturability is
thus established by the directly measured coverage dynamic range, not inferred from
$Da_h$ or $R_{\mathrm{sup}}$.

\subsection{Mesh and grid independence}
A boundary layer is placed on the substrate (first layer $0.5~\mu$m, 10 layers,
stretch 1.2); the bulk maximum element size is $40~\mu$m, all scaled by a uniform
refinement factor $m_f$. The reference case is solved manually at successive $m_f$.
Table~\ref{tab:meshconv} lists three refinement levels. The fitted integral label
$\thetabar$ is converged to $\pm0.1\%$ (it varies only within $0.7742$--$0.7759$
across $m_f=2,4,8$), whereas the near-wall concentration $\langle c\rangle/C_0$ does
\emph{not} converge---even as a surface average it swings non-monotonically
($0.0148\!\to\!0.0123\!\to\!0.0206$, a $67\%$ jump from $m_f=4$ to $8$), because it
samples the extremely steep leading-edge gradient in the strong-adsorption regime.
The production mesh is $m_f=2$ ($\approx3.5\times10^4$ elements).

\begin{table}[t]\centering
\caption{Mesh convergence at the reference case ($\vsub=0.05$, $\Ucur=0.5$, $T=300$~K).
The integral label $\thetabar$ converges; the near-wall concentration
$\langle c\rangle/C_0$ does not.}
\label{tab:meshconv}
\begin{tabular}{@{}rrrr@{}}
\toprule
$m_f$ & elements & $\thetabar$ & $\langle c\rangle/C_0$ \\
\midrule
2 & 35{,}393  & 0.77589 & 0.0148 \\
4 & 104{,}406 & 0.77416 & 0.0123 \\
8 & 331{,}069 & 0.77512 & 0.0206 \\
\bottomrule
\end{tabular}
\end{table}

This near-wall pathology does \emph{not}, however, contaminate the kinetic inversion.
The adsorption energy $\Eads$ is recovered from the \emph{temperature dependence} of the
converged integral $\thetabar$, not from the amplitude of $\langle c\rangle/C_0$; and
the product $\kads\cwall$ is in any case structurally non-identifiable (Section~6.5),
so the mesh sensitivity of $\cwall$'s scale only translates the solution along that
already-degenerate direction without moving the identifiable $\Eads$ or the degeneracy
slope. Two facts make this concrete rather than merely plausible. First, because
temperature enters only through $\kdes$ and the flow is decoupled from temperature (the
measured wall shear rates are identical across all temperatures, Section~4.2), any
residual mesh bias in $\thetabar$ is \emph{common-mode across temperatures} and
therefore largely cancels in the Arrhenius slope $d\ln\kdes/d(1/T)$ from which $\Eads$
is read; the $\pm0.1\%$ spread of $\thetabar$ across $m_f=2,4,8$ (Table~\ref{tab:meshconv}) thus
propagates to a negligible shift in $\Eads$. Second, this is confirmed \emph{empirically}
by the independent coarse-grid experiment of Section~7.4: regenerating the dataset on a
much coarser mesh (maximum element $80~\mu$m, a $>2\times$ coarsening that exercises
exactly the near-wall pathology reported here) shifts the empirical degeneracy slope only
to $0.063$ (versus $0.0647$ on the fine grid) and the conditional $\Eads$ to $0.777$~eV
(versus $0.776$)---a $\sim1$~meV change---so the $\Eads$ profile-likelihood minimum is
mesh-stable to well within its own uncertainty. The single-point wall concentration does
not converge under refinement (extremely steep near-wall gradient in the
strong-adsorption regime), so near-wall concentration observations use the stable
surface/line average $\langle c\rangle/C_0$ only as an auxiliary supervision label
anchoring the \emph{magnitude} of $C_s^*$ (validated quantitatively in
Sections~6.1 and~6.6); the integral
quantities $\thetabar$ and $J_{\mathrm{net}}$ converge well.

\subsection{Case scan and solving}
The main scan takes $\vsub\in\{0.05,0.1,0.2,0.3,0.5,0.8,1.2,2.0\}$~m/s (8 values)
and $\Ucur\in\{0.5,1,2,4,8,16\}$~m/s (6 values), a full cross of 48 cases. The
scan ranges are chosen by physical purpose rather than arbitrarily: the
lower $\vsub$ bound ensures a saturable end, the upper bound covers high-throughput
roll-to-roll/rotary operation, and the curtain range covers from ``almost no
isolation'' to ``strong curtain strongly suppressing near-wall concentration'';
together they guarantee the full saturated--transition--undersaturated dynamic
range (about four orders of magnitude) required for kinetic invertibility. With
the segregated-step damping lowered from $0.7$ to $0.5$ and the maximum iterations
raised from $300$ to $500$, all 48 cases converge. Each case exports the
substrate-averaged coverage $\thetabar$ (primary label), the exit/peak coverage
(diagnostics), the surface-averaged near-wall concentration $C_{s,A}^{mid}$
(auxiliary supervision), and conservation diagnostics.

\subsection{Dataset and split}
The dataset has 48 rows, each recording $(\vsub,\Ucur)$, their normalisations
$v^*=\vsub/2.0$ and $U^*=\Ucur/16$, and the derived scalars. Here the divisors
$2.0$ and $16$ are simply the upper bounds of the respective parameter scans
(map-to-$[0,1]$ feature scaling); they are not physical constants. It is stratified into
$30$ train / $9$ validation / $9$ test, with near-saturated cases placed mainly in
the training set so that the desorption information can be learned. Training and
validation labels carry $\sigma=5\%$ relative noise; the test set is clean. The
substrate-averaged coverage $\thetabar$ spans about four orders of magnitude (from
the saturated end $0.78$ to the strong-curtain/fast-substrate end $\sim10^{-4}$),
with the curtain velocity as the dominant control, forming a clear operating window
in the $(\vsub,\Ucur)$ plane.

\section{Physics--chemistry-informed neural network (PCINN)}

\subsection{Overall architecture}
PCINN couples two branches in series. The physics branch (a neural network)
approximates the operating-condition$\,\to\,$effective near-wall concentration
closure; the chemistry branch (a hard-coded Langmuir model with trainable
kinetics) integrates coverage along the substrate trajectory. The data-driven
freedom is compressed to a single scalar $C_s^*$, with everything else carried by
known physics:
\begin{equation}
(v^*,U^*)\;\xrightarrow{\text{physics branch}}\;C_s^*
\;\xrightarrow{\cwall=C_s^*C_0}\;\text{chemistry branch (trajectory integration)}
\;\to\;\thetabar .
\end{equation}

\begin{figure}[t]\centering
\includegraphics[width=0.92\textwidth]{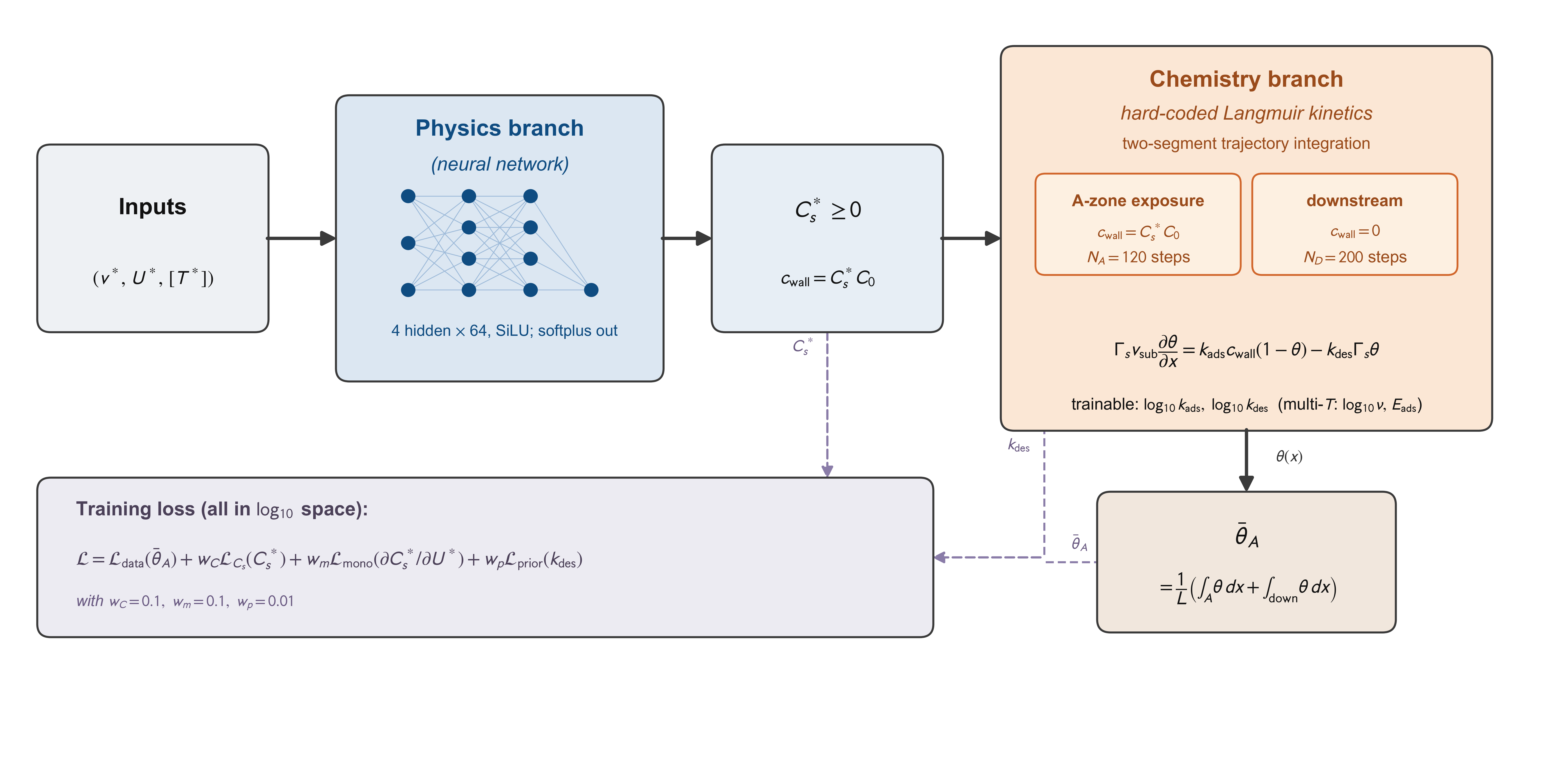}
\caption{PCINN architecture. A physics branch (small MLP) maps the operating
condition $(v^*,U^*[,T^*])$ to an effective near-wall concentration $C_s^*$; a
hard-coded chemistry layer with trainable kinetics integrates coverage along the
substrate trajectory in two segments (A-zone exposure and downstream desorption) to
yield $\thetabar$. The data-driven freedom is compressed to a single scalar.}
\label{fig:arch}
\end{figure}

\subsection{Physics branch}
A fully connected network with inputs $(v^*,U^*)$ (plus $T^*$ for the multi-temperature
case), four hidden layers of 64 units, SiLU activation, and a softplus output
ensuring $C_s^*\ge0$. The output is not capped at $C_0$, allowing the network to
learn an effective driving concentration; the learned effective $\cwall$ is about
$5.5\times$ the surface-averaged $C_{s,A}^{mid}$, which is physically reasonable
(Section~6.1).

\subsection{Chemistry branch and two-segment trajectory coupling}
The chemistry branch reuses the same spatial advection equation, with the near-wall
concentration replaced by the physics-branch output, integrated forward along the
substrate. Kinetic parameters are parameterised in $\log_{10}$ space. To make the
output consistent with the primary label $\thetabar$ (whole-substrate average), the
integration is split into two segments: an A-zone exposure segment (length
$L_A=w_A=5$~mm, $\cwall=C_s^*C_0$) and a downstream desorption segment
(length $L_{\mathrm{down}}=22.5$~mm, $\cwall=0$), with the substrate average
\begin{equation}
\thetabar=\frac{1}{L}\Big(\int_A\theta\,dx+\int_{\mathrm{down}}\theta\,dx\Big),
\end{equation}
discretised by forward Euler ($N_A=120$, $N_{\mathrm{down}}=200$ steps),
fully differentiable.

The trajectory integrator is an ordinary differential equation in the coverage $\theta$
(not an advection PDE), so it is free of numerical diffusion; within each segment the
kinetics are \emph{linear} in $\theta$ ($d\theta/dx=A-(A+B)\theta$ in the A zone with
$A=\kads\cwall/(\Gamma_s\vsub)$, $B=\kdes/\vsub$, and $d\theta/dx=-B\theta$ downstream),
which admits an exact solution against which the forward-Euler discretisation is checked.
The explicit scheme is unconditionally usable in this regime: at the stiffest operating
point the dimensionless step parameter $(A+B)\,dx\approx0.04\ll2$, far from the
stability limit. Step-independence is verified against this analytic reference
(Table~\ref{tab:euler}): at the production resolution the relative error in $\thetabar$
is $0.17\%$ (median over the operating window) and at most $1.4\%$ (at the extreme
low-coverage corner $\vsub=0.05$, high $T$, $\theta\sim5\times10^{-4}$), both well below
the $\sigma=5\%$ label noise; the error halves under each doubling of the step count
(clean first-order convergence), confirming that $N_A=120$, $N_{\mathrm{down}}=200$ lies
on the converged plateau.

\begin{table}[t]\centering
\caption{Forward-Euler step independence. Relative error in $\thetabar$ against the
exact per-segment linear-ODE solution, at the worst-case (extreme low-coverage) operating
point. The production resolution ($N_A=120$, $N_{\mathrm{down}}=200$) already sits well
below the $\sigma=5\%$ label noise; the error halves under each step doubling.}
\label{tab:euler}
\small
\begin{tabular}{lc}
\toprule
$N_A$ / $N_{\mathrm{down}}$ & worst-case rel.\ error in $\thetabar$ \\
\midrule
$30$ / $50$ & $5.7\%$ \\
$60$ / $100$ & $2.8\%$ \\
$120$ / $200$ (production) & $1.4\%$ \\
$240$ / $400$ & $0.71\%$ \\
$480$ / $800$ & $0.35\%$ \\
\bottomrule
\end{tabular}
\end{table}

\begin{figure}[t]\centering
\includegraphics[width=0.95\textwidth]{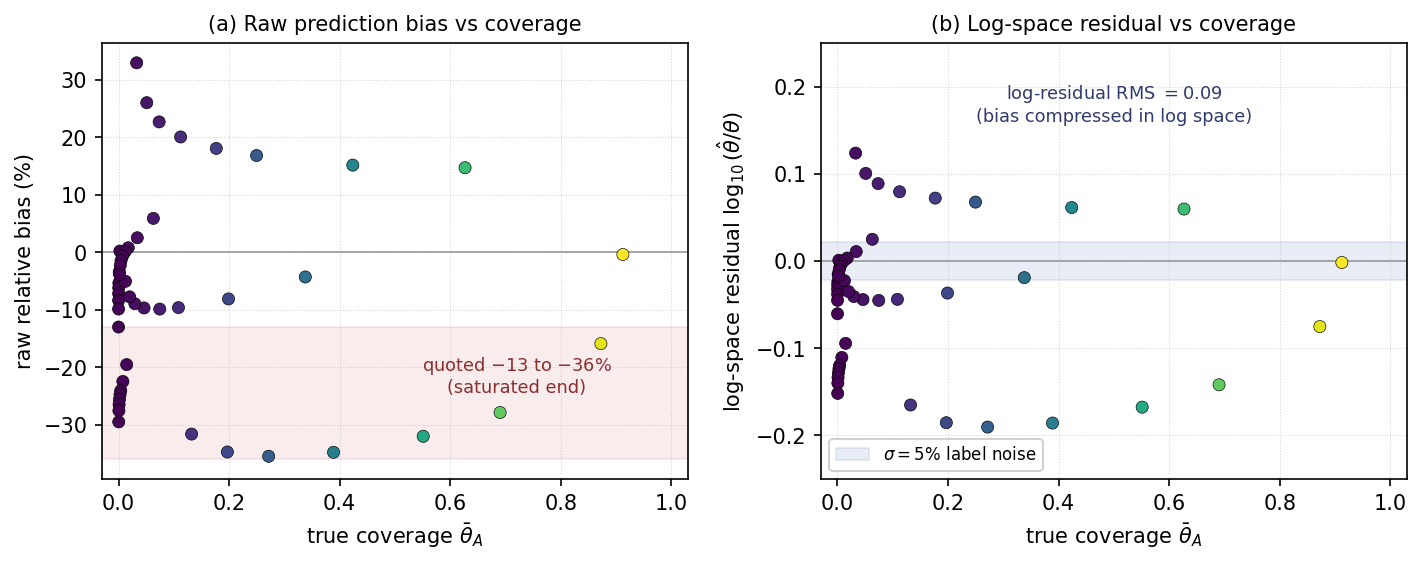}
\caption{Model-form bias versus coverage. (a) Raw relative prediction bias
$(\hat\theta-\theta)/\theta$: a systematic under-prediction of $13$--$36\%$ appears only
at the saturated (high-coverage) end, the expressivity limit of the single-scalar
near-wall closure. (b) The same residuals in log space
$\log_{10}(\hat\theta/\theta)$ stay within the $\sigma=5\%$ label-noise band across four
decades of coverage (RMS $0.09$); the bias is therefore real but log-compressed, and
perturbs the log-space kinetic inversion by only $\approx0.016$~eV
($\Eads$), leaving the identifiability conclusions intact.}
\label{fig:biascov}
\end{figure}

A known model-form bias follows from approximating the entire
A-zone exposure by a single effective $\cwall$ and the downstream by pure
desorption: at the saturated end the high-coverage cases are systematically
under-estimated by $\sim13\%$--$36\%$. Physically, this is the expressivity limit of the
single-scalar near-wall-concentration closure in the strong-depletion / near-saturation
regime, where one effective $\cwall$ cannot reproduce the full streamwise concentration
profile. This bias is compressed in log space
($\Rlog$ remains $\sim0.998$) and is the main cause of the $R^2_{\mathrm{raw}}
\approx0.989$ deduction; its contribution to the multi-temperature $\Eads$
inversion is later quantified at $\approx0.016$~eV (Section~7), so it does \emph{not}
threaten the identifiability conclusions. A bias-versus-coverage diagnostic
(Fig.~\ref{fig:biascov}) makes this explicit: the raw relative bias reaches tens of
percent only at the highest coverage, yet the log-space residual stays within the
$\sigma=5\%$ label-noise band across all four decades of coverage (RMS $0.09$).

\subsection{Loss function}
The loss combines the data term, the near-wall concentration supervision, a
monotonicity prior, and a weak kinetic prior:
\begin{equation}
\mathcal{L}=\mathcal{L}_{\mathrm{data}}+w_C\mathcal{L}_{C_s}
+w_m\mathcal{L}_{\mathrm{mono}}+w_p\mathcal{L}_{\mathrm{prior}},
\qquad w_C=0.1,\ w_m=0.1,\ w_p=0.01,
\end{equation}
with all terms in $\log_{10}$ space. The data term in log space is essential
because $\theta$ spans four orders of magnitude (a raw-value MSE is dominated by
large values and collapses on the low-coverage cases, $\Rlog\approx0.68$, versus
$\sim0.998$ in log space). The concentration supervision $\mathcal{L}_{C_s}$ uses
the surface/line average; the monotonicity prior penalises
$\partial C_s^*/\partial U^*>0$ (a stronger curtain lowers near-wall
concentration); the weak kinetic prior $(\log_{10}\kdes-\log_{10}\kdes^{\mathrm{prior}})^2$
is used at a single temperature and removed for the multi-temperature inversion (where the
Arrhenius relation across temperatures provides the constraint). The Adam optimiser
($10^{-3}$) with ReduceLROnPlateau is used, with early stopping on the validation
log data-loss; headline results are reported as mean$\pm$s.d.\ over 8 random seeds,
each perturbing both the initialisation and the label-noise realisation.

\section{Results and discussion}

\subsection{Surrogate accuracy}
Using only 30 training cases, PCINN reproduces $\thetabar$ across about four orders
of magnitude. Over 8 random seeds (each perturbing both initialisation and label
noise), the test $\Rlog=0.9975\pm0.0005$ and $R^2_{\mathrm{raw}}=0.989\pm0.003$;
the train/val/test $\Rlog$ are $0.991/0.996/0.9975$ (cross-seed s.d.\ $\le0.001$).
The test $\Rlog$ is slightly above the training value on every seed---not because
there is no over-fitting, but because the worst-fit high-coverage cases happen to
fall mostly in the training set; the robust statement is ``no appreciable
over-fitting gap''.

The effective near-wall concentration learned by the physics branch agrees in
magnitude with the COMSOL surface-averaged $C_{s,A}^{mid}$ but is systematically
about $5.5\times$ higher (range $3.3$--$7.9\times$). This is \emph{not} a
supervision failure but reflects the network learning the physically correct
effective driving concentration, which can be verified quantitatively from the
COMSOL field. At the saturated case ($\vsub=0.05$, $\Ucur=0.5$, $T=300$~K), the
near-wall $c/C_0$ along the A zone decays sharply from the leading edge: $0.124$
at $x=2.5$~mm, $0.059$ at the quarter point, $0.015$ at the midpoint, $0.0011$ at
the three-quarter point, $0.0003$ at the trailing edge---about $400\times$ from
leading edge to end (the precursor is rapidly consumed on entering the A zone).
The leading-edge concentration $0.124$ nearly coincides with the learned
$C_s^*\approx0.021\times5.5\approx0.115$, whereas the surface average
$C_{s,A}^{mid}\approx0.021$ mixes the high leading-edge value with the near-zero
trailing values and severely under-estimates the effective concentration. The
auxiliary supervision $\mathcal{L}_{C_s}$ therefore \emph{anchors $C_s^*$ at the
correct magnitude} (preventing $C_s^*$ from drifting freely under the
$\kads\!\cdot\!\cwall$ degeneracy), while within that constraint the network learns
the effective leading-edge concentration; the two are not in conflict.
The spread of the ratio ($3.3$--$7.9\times$) is itself systematic rather than scatter:
it tracks how sharply the near-wall concentration is depleted along the A zone, set by
the balance between adsorption consumption and convective/diffusive resupply. Strongly
consuming conditions (low $\vsub$ or near-saturation) produce a steeply peaked profile
and a large leading-edge-to-average ratio, whereas weakly consuming conditions (high
$\vsub$, a strong curtain, or high $T$ with fast desorption) flatten the profile and
push the ratio toward unity. Because the auxiliary loss anchors only the order of
magnitude while the network learns this condition-dependent effective concentration
within that constraint, the spread does not degrade the anchoring: the surrogate
accuracy and, in particular, the $\Eads$ inversion (read from the temperature slope, not
the absolute $\cwall$ scale) are unaffected.
This also
explains the $\sim7\times$ off-truth $\kads$ minimum in the profile likelihood of
Section~6.5: supervision anchors only the magnitude, not the absolute value, so the
residual concentration-scale freedom can still be absorbed by $\kads$.

\begin{figure}[t]\centering
\includegraphics[width=0.58\textwidth]{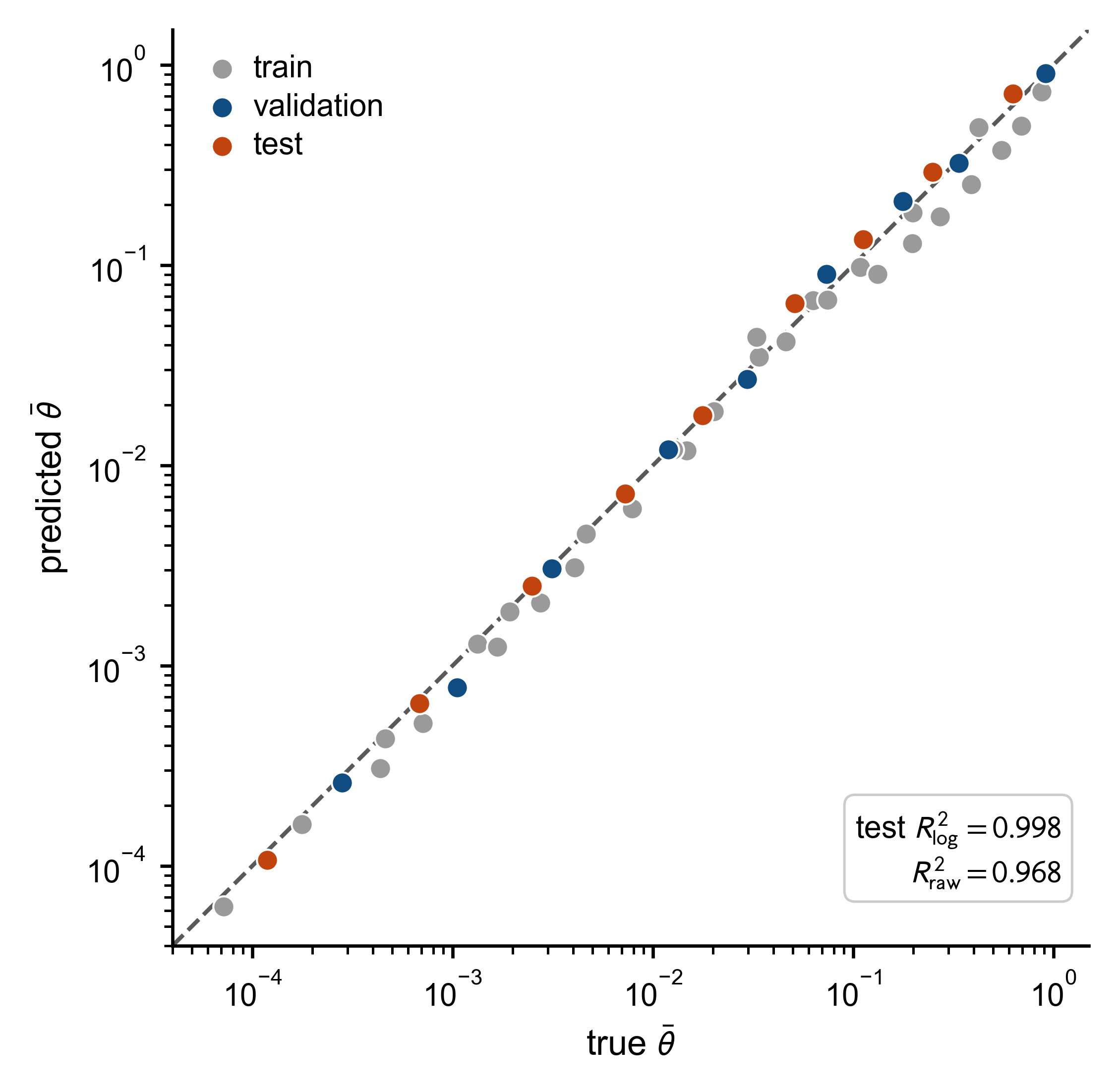}
\caption{Surrogate accuracy (representative run). Predicted versus true
$\thetabar$ across about four orders of magnitude on the clean test set; PCINN reaches
test $\Rlog=0.998$ using only 30 training cases.}
\label{fig:parity}
\end{figure}

\subsection{Sample efficiency: physics versus a data-only baseline}
The comparisons in this and the following two subsections are not an end in themselves:
their purpose is to establish that the embedded physics buys \emph{invertibility and a
trustworthy identifiability analysis}, not merely surrogate accuracy---so the value of
the physics constraint is relocated from ``fitting accuracy'' to ``invertibility $+$
extrapolation $+$ identifiability-boundary characterisation'', which the rest of the
paper develops. To quantify the gain from the physics constraint, we compare PCINN with a
structure-free multilayer perceptron (MLP). Both comparisons (sample efficiency
and extrapolation) are performed in the \emph{primary caliber} ($\thetabar$,
two-segment), consistent with the main results; the MLP directly regresses
$\log_{10}\thetabar$ from $(v^*,U^*)$ with comparable capacity and identical loss,
optimiser and early-stopping. At the sparsest $n=6$, PCINN attains
$\Rlog=0.915\pm0.005$ with high seed consistency, whereas the MLP gives only
$0.78\pm0.13$ with an order of magnitude larger variance ($0.65$--$0.92$); PCINN
saturates at $0.997\pm0.001$ by $n\approx15$, while the MLP catches up to $0.990$
only at $n=30$.

To calibrate the magnitude of the ``physics gain'', we add a Gaussian-process (GP)
regression baseline (anisotropic Mat\'ern kernel, $(v^*,U^*)\to\log_{10}\thetabar$),
a comparison deliberately favourable to the baseline since a 2-D dense sampling is
the GP's home ground (Table~\ref{tab:gp}). At full data the GP reaches
$\Rlog=0.983$ ($R^2_{\mathrm{raw}}=0.936$), below PCINN's $0.998$ but already high,
confirming that a pure interpolator suffices for the surrogate task under dense
sampling. In sample efficiency, however, the gap widens sharply at the sparse end:
at $n=6$ the GP gives only $0.58\pm0.42$ (nearly unusable) versus PCINN's
$0.92\pm0.01$, and stabilises above $0.96$ only by $n\approx18$. This clarifies the
true positioning of the PCINN gain: when data are abundant the GP interpolation
accuracy approaches PCINN, so the value of the physics constraint is \emph{not}
mainly in interpolation accuracy; its irreplaceable advantages are (i) stable high
accuracy under data scarcity, (ii) invertible physical parameters ($\Eads$,
$\kdes$, which a GP cannot provide), and (iii) extrapolation along the residence-time
axis and identifiability analysis (Sections~6.4, 6.5 and~7). The comparison with
the GP thus does not weaken the contribution but relocates it from ``surrogate
accuracy'' to ``invertibility $+$ extrapolation $+$ identifiability-boundary
characterisation''.

\begin{table}[t]\centering
\caption{GP baseline versus PCINN (primary caliber $\thetabar$).}
\label{tab:gp}
\small
\begin{tabular}{lcc}
\toprule
training cases $n$ & GP $\Rlog$ & PCINN $\Rlog$ \\
\midrule
6   & $0.58\pm0.42$ & $0.92$ \\
12  & $0.92\pm0.12$ & $0.95$ \\
18  & $0.96\pm0.02$ & $0.97$ \\
24  & $0.98\pm0.01$ & $0.98$ \\
30  & $0.98\pm0.01$ & $0.985$ \\
full (test) & $0.983$ & $0.998$ \\
\bottomrule
\end{tabular}
\end{table}

\begin{figure}[t]\centering
\includegraphics[width=0.62\textwidth]{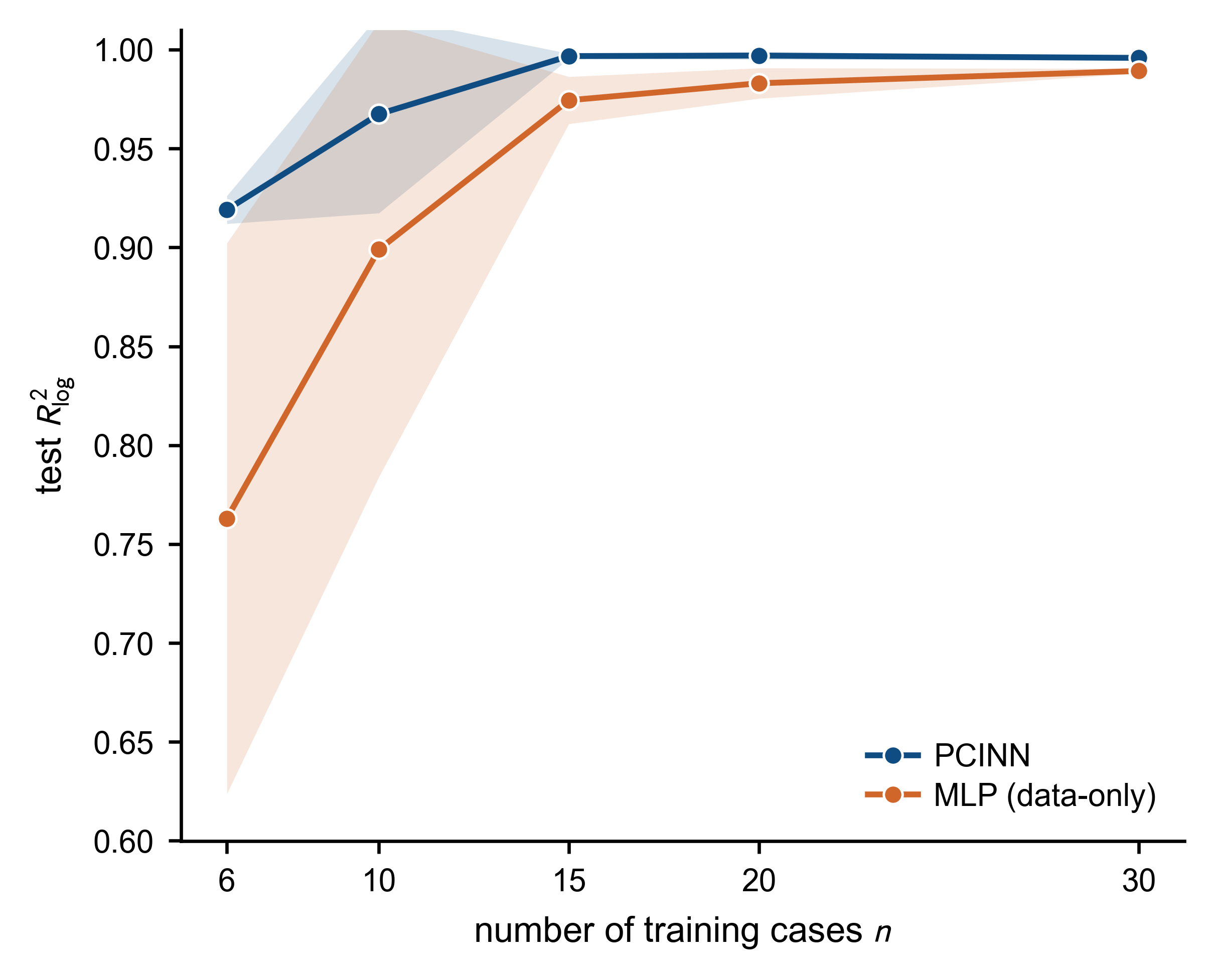}
\caption{Sample efficiency (primary caliber $\thetabar$, two-segment). PCINN attains
$\Rlog\approx0.92$ already at $n=6$ training cases with small seed variance, whereas a
structure-free MLP needs $\sim30$ cases to catch up and has an order-of-magnitude
larger variance at the sparse end.}
\label{fig:sampleeff}
\end{figure}

\subsection{Comparison with an analytic baseline}
We compare against the Yanguas-Gil--Elam plug-flow self-limiting model (which
assumes sufficient supply $c=C_0$ and integrates along residence time, depending
only on $\vsub$). Of the 48 cases, only 5 (slow-substrate/low-curtain saturated
corner) fall within $2\times$ of the analytic value, with a median over-prediction
of $1.55$ decades and up to $3.94$ decades at high curtain---clearly delineating the
``analytic-fails, PCINN/resolved-transport-required'' regime. (This comparison uses
$\theta_{\mathrm{exit}}$ since the analytic expression only yields an exit
coverage, as noted.) We stress that this is \emph{not} a fair accuracy contest and we
do not use it to inflate PCINN: the plug-flow model fails here \emph{by construction},
because its sufficient-supply assumption ($c=C_0$) is exactly what the gas curtain
violates, so it is structurally incapable of representing the curtain-dominated bulk of
the window. The comparison should therefore be read as an \emph{operating-domain
partition}---identifying where a cheap analytic estimate remains usable (the
supply-rich saturated corner) versus where resolved transport is mandatory---rather
than as evidence that PCINN is "more accurate" than a model built for a different
regime.

\begin{figure}[t]\centering
\includegraphics[width=0.98\textwidth]{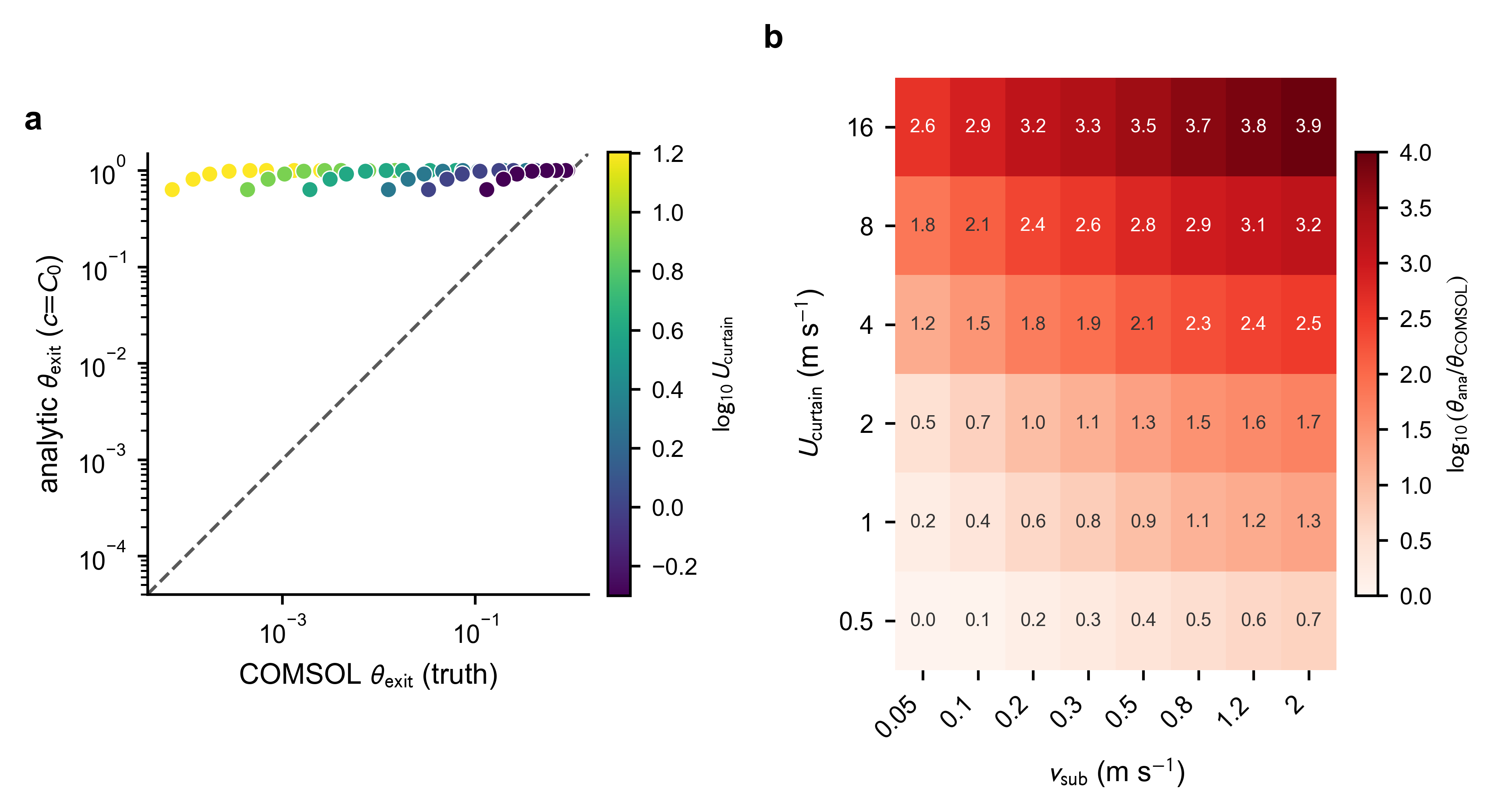}
\caption{Analytic baseline (Yanguas-Gil--Elam plug-flow self-limiting model). Only the
slow-substrate/low-curtain saturated corner falls within $2\times$ of the analytic
value; the median over-prediction is $1.55$ decades (up to $3.94$ at high curtain),
delineating the regime where resolved transport (PCINN) is required.}
\label{fig:analytic}
\end{figure}

\subsection{Extrapolation: the action boundary of the physics constraint}
Extrapolating along the substrate velocity $\vsub$ (train on $\vsub\le0.5$, predict
$\vsub\ge0.8$), PCINN stays robustly positive across all $8$ seeds
($\Rlog=0.90\pm0.13$, range $0.61$--$0.998$), whereas the MLP is unstable
($-0.16\pm1.01$, range $-2.01$ to $0.95$, an order-of-magnitude larger spread and
frequently negative). The reason is that the coverage dependence on $\vsub$
enters through the residence time $\tau=w_A/\vsub$, which is \emph{known physics} in
the chemistry-branch integration, while the effective near-wall concentration grows
only mildly with $\vsub$. Conversely, extrapolating along the curtain velocity
$\Ucur$ (train on $\Ucur\le2$, predict $\Ucur\ge4$), both models fail and are
statistically indistinguishable ($8$-seed mean$\pm$s.d.\ PCINN $-7.9\pm5.1$, MLP
$-7.8\pm3.3$), because $\Ucur$ drives precisely the
operating-condition$\,\to\,$$C_s^*$ mapping that is learned by the network and
unconstrained by the chemistry layer. Thus the physics constraint provides
extrapolation gains \emph{only} along the structurally known (residence-time) axis,
and confers no advantage along the data-driven transport axis---a deliberately
honest negative control.

\begin{figure}[t]\centering
\includegraphics[width=0.62\textwidth]{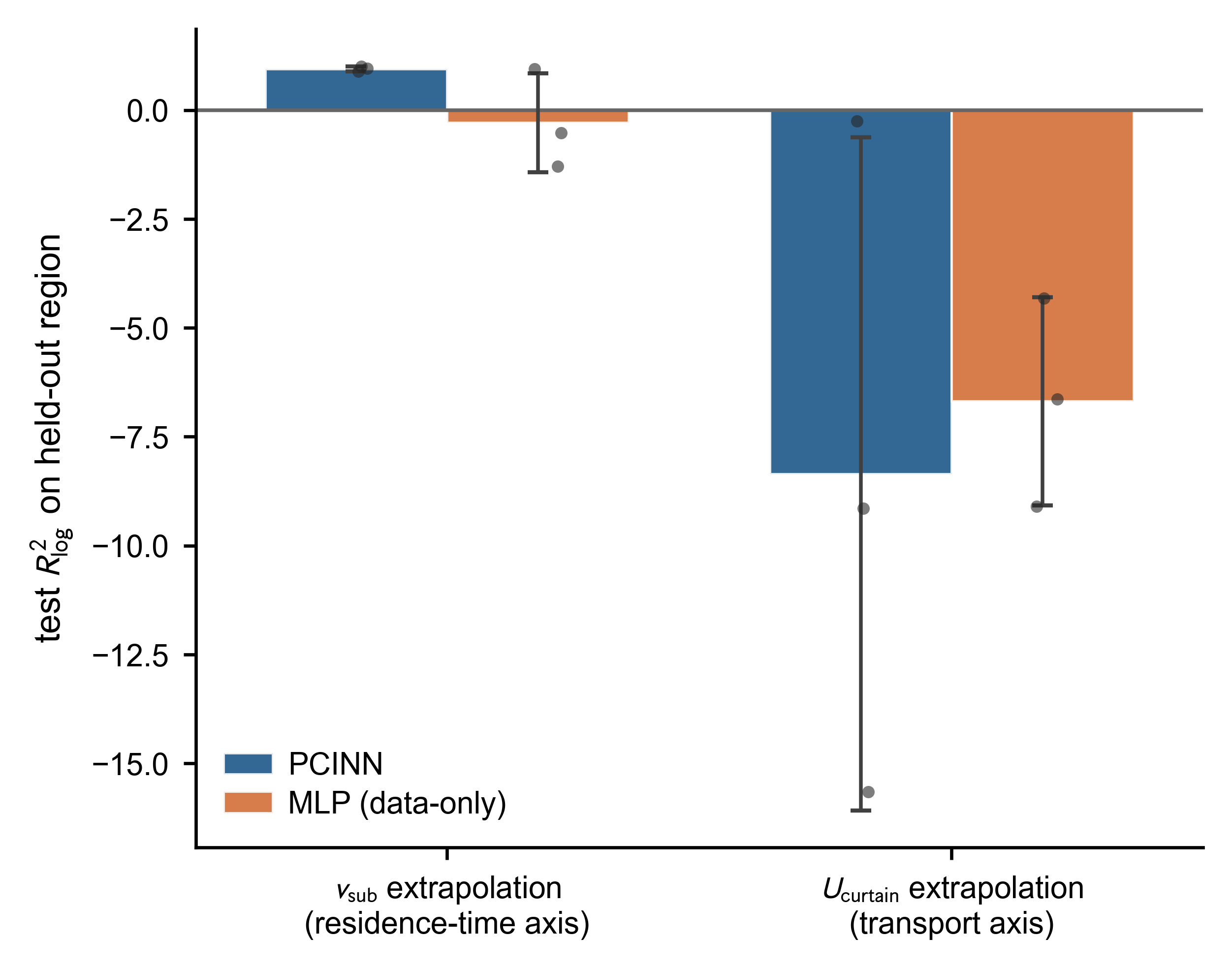}
\caption{Extrapolation to held-out operating regions ($8$ seeds, mean$\pm$s.d.; dots
are individual seeds). Along the substrate velocity $\vsub$ (residence time, carried
by the chemistry branch) PCINN stays robustly positive ($\Rlog=0.90\pm0.13$, all
seeds $>0$) while the MLP is unstable; along the curtain velocity $\Ucur$ (the
data-driven transport axis) both fail and are indistinguishable. The physics
constraint helps only along the structurally known axis.
($R^2_{\log}<0$ means the prediction is worse than a constant-mean baseline,
i.e.\ a failed extrapolation.)}
\label{fig:extrap}
\end{figure}

\subsection{Kinetic inversion and identifiability}
\textbf{Identifiable.} $\Eads$ and $\kdes$ are robustly identifiable: over 8 seeds,
$\Eads=0.773\pm0.001$~eV (truth $0.774$) and $\kdes=1.04\pm0.03$~s$^{-1}$ (truth
$1.0$). Note that $\Eads$ and $\kdes$ are two expressions of the same identifiable
quantity ($\Eads=k_BT\ln(\nu/\kdes)$ at fixed $\nu,T$); the narrower-looking $\Eads$
interval is purely log compression. $\kdes$ can be anchored because the dataset
contains near-saturated cases (where the desorption term has appreciable
magnitude).

\textbf{Not separately identifiable.} $\kads$: the data constrain only the
adsorption-flux product $\kads\cwall$, so $\kads$ is reported as an effective value
(8 seeds $0.0116\pm0.0005$). The small cross-seed s.d.\ is not evidence of
identifiability (all runs share the initial value); the non-identifiability is
established by three lines of evidence.

\textbf{Three mechanisms.} (i) \emph{Structural (Fisher).} With the learned
near-wall concentration closure fixed, the Fisher information matrix of
$(\log_{10}\kads,\log_{10}\kdes)$ has condition number $\approx1.4\times10^2$; once
the concentration scale of the closure is allowed to vary, the condition number
jumps to $\approx1.25\times10^9$ with the flattest eigendirection $(+1,-1)/\sqrt2$,
i.e.\ information vanishes along the $\kads\cwall$-invariant direction. The Fisher
matrix is constructed as follows: with the parameter vector $\boldsymbol{\beta}$ and
the model log-coverage prediction $f_i(\boldsymbol{\beta})=\log_{10}\bar{\theta}_{A,i}$
at each case $i$, we take the autodiff gradients $\partial f_i/\partial\beta_j$ at
the converged solution and assemble the Gauss--Newton approximation
$F_{jk}=\sigma^{-2}\sum_i(\partial f_i/\partial\beta_j)(\partial f_i/\partial\beta_k)$.
Here $F$ is taken from the \emph{data term only}, excluding the auxiliary
supervision and regularisers, in order to isolate the constraint of the data itself.
(ii) \emph{Initial-value drift.} Varying the $\kads$ initial value
($0.001/0.01/0.1$) drives the converged value monotonically ($0.0104/0.0376/0.0585$),
not to the truth. (iii) \emph{Profile likelihood.} The $\kads$ profile has a clear
minimum but offset from the truth by $\sim7\times$ ($\kads\approx0.07$ vs $0.01$),
i.e.\ the fit actively prefers a wrong $\kads$; the $\Eads$ profile minimum lies near
the truth. The three differ in whether the minimum sits at the truth, not in
``flat versus sharp''.

\textbf{Deepened structural analysis.} Beyond the condition number, we examine the
full eigenstructure of the closure-free Fisher matrix in
$(\log_{10}\kads,\log_{10}\kdes,\log_{10}c\text{-scale})$. Its eigenvalues span
$\lambda_1\approx3.3\times10^4$, $\lambda_2\approx1.1\times10^2$, and $\lambda_3$ at
the double-precision floor ($|\lambda_3|/\lambda_1\approx2\times10^{-16}$, an
\emph{exact structural zero} contaminated only by round-off), with null eigenvector
$(+1,0,-1)/\sqrt2$---the $\kads\cwall$-invariant direction. This structural
(model-level) non-identifiability of $\kads$ admits an analytic proof: the adsorption
flux enters Eq.~(2) only as the product $\kads\cwall$, so the reparameterisation
$\kads\!\to\!\alpha\kads$, $\cwall\!\to\!\cwall/\alpha$ leaves \emph{every} prediction
unchanged; the likelihood is exactly flat along it and the corresponding Fisher
eigenvalue is identically zero. Because the matrix is structurally singular we do not
quote a finite condition number for it (the $1.25\times10^9$ above is the value when
the concentration scale is only \emph{partially} constrained by the supervision,
i.e.\ a practical-identifiability estimate). Displacing the parameters by $\pm10\%$
along this null direction changes the model output by less than the label noise (a
direct visualisation of the sloppy manifold), whereas the same displacement along
$\lambda_1$ changes it by orders of magnitude; the singular-value spectrum of the
prediction-sensitivity matrix $\partial f_i/\partial\beta_j$ mirrors this hierarchy.
We therefore distinguish \emph{structural} non-identifiability ($\kads$ alone, exactly
unrecoverable for any data) from \emph{practical} non-identifiability ($\nu$ at
multiple temperatures, recoverable only to a sub-decade interval ($\chi^2_1$ $95\%$
width $\approx0.6$ decade) given finite, noisy data; Section~7.2).

\begin{figure}[t]\centering
\includegraphics[width=0.92\textwidth]{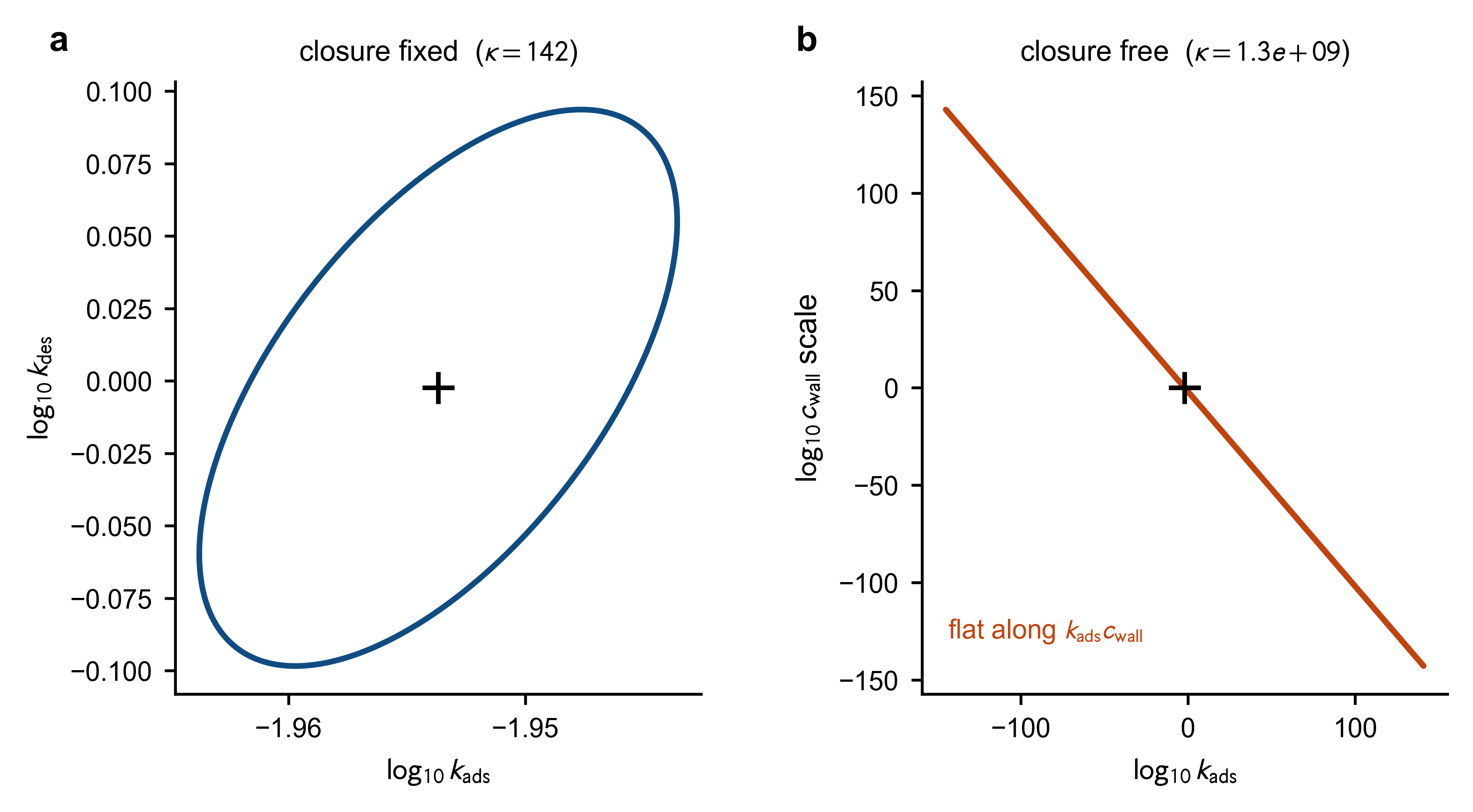}\\[6pt]
\includegraphics[width=0.92\textwidth]{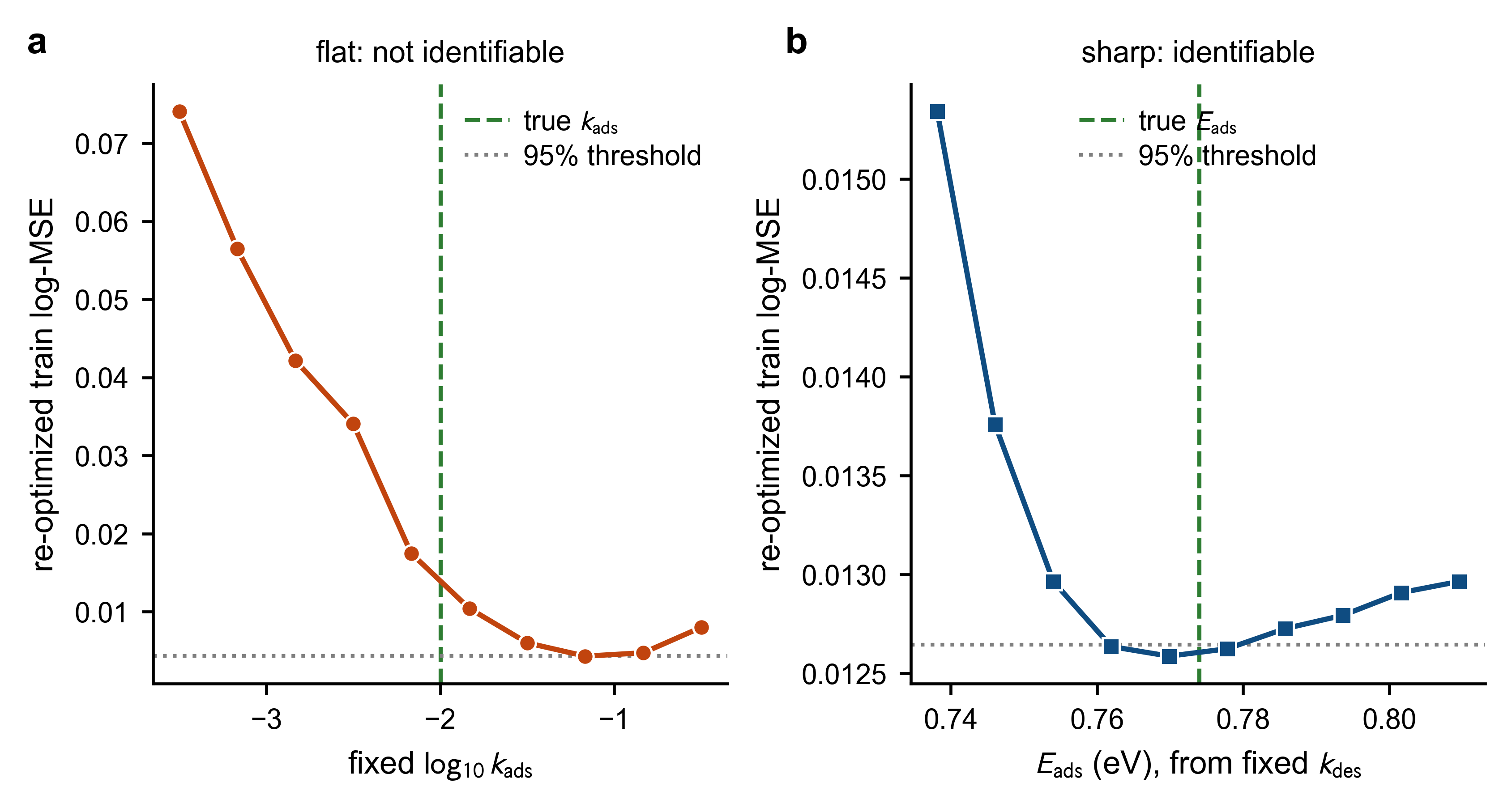}
\caption{Single-temperature identifiability. Top: Fisher confidence ellipses---once
the concentration scale is freed, the likelihood is nearly flat along the
$\kads\cwall$-invariant direction. Bottom: profile likelihood---the $\kads$ profile has
a clear minimum offset from the truth by $\sim7\times$ (minimum off-true), whereas the
$\Eads$ profile minimum sits at the truth.}
\label{fig:fisher}
\end{figure}

\begin{figure}[t]\centering
\includegraphics[width=0.49\textwidth]{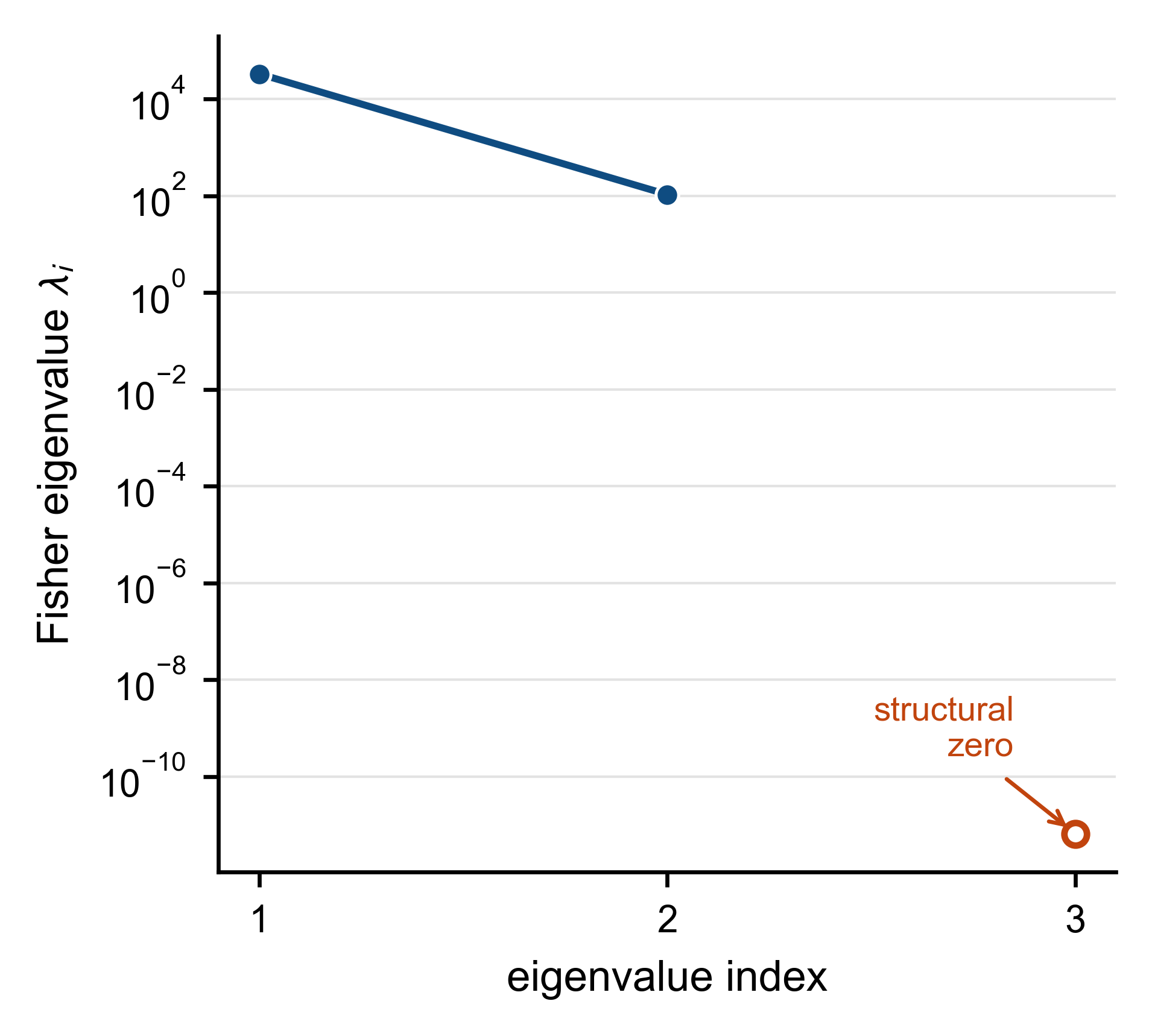}\hfill
\includegraphics[width=0.49\textwidth]{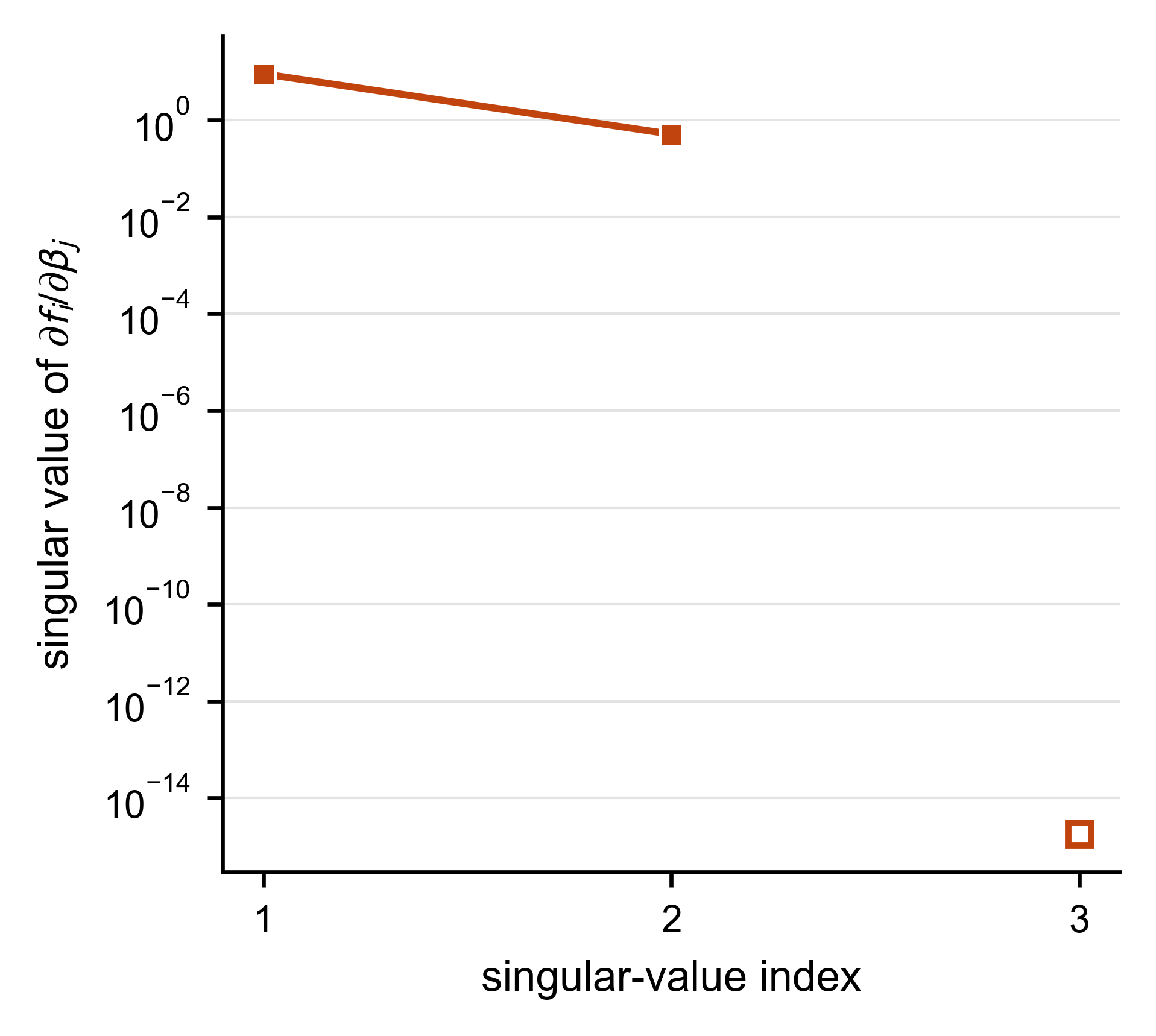}\\[4pt]
\includegraphics[width=0.6\textwidth]{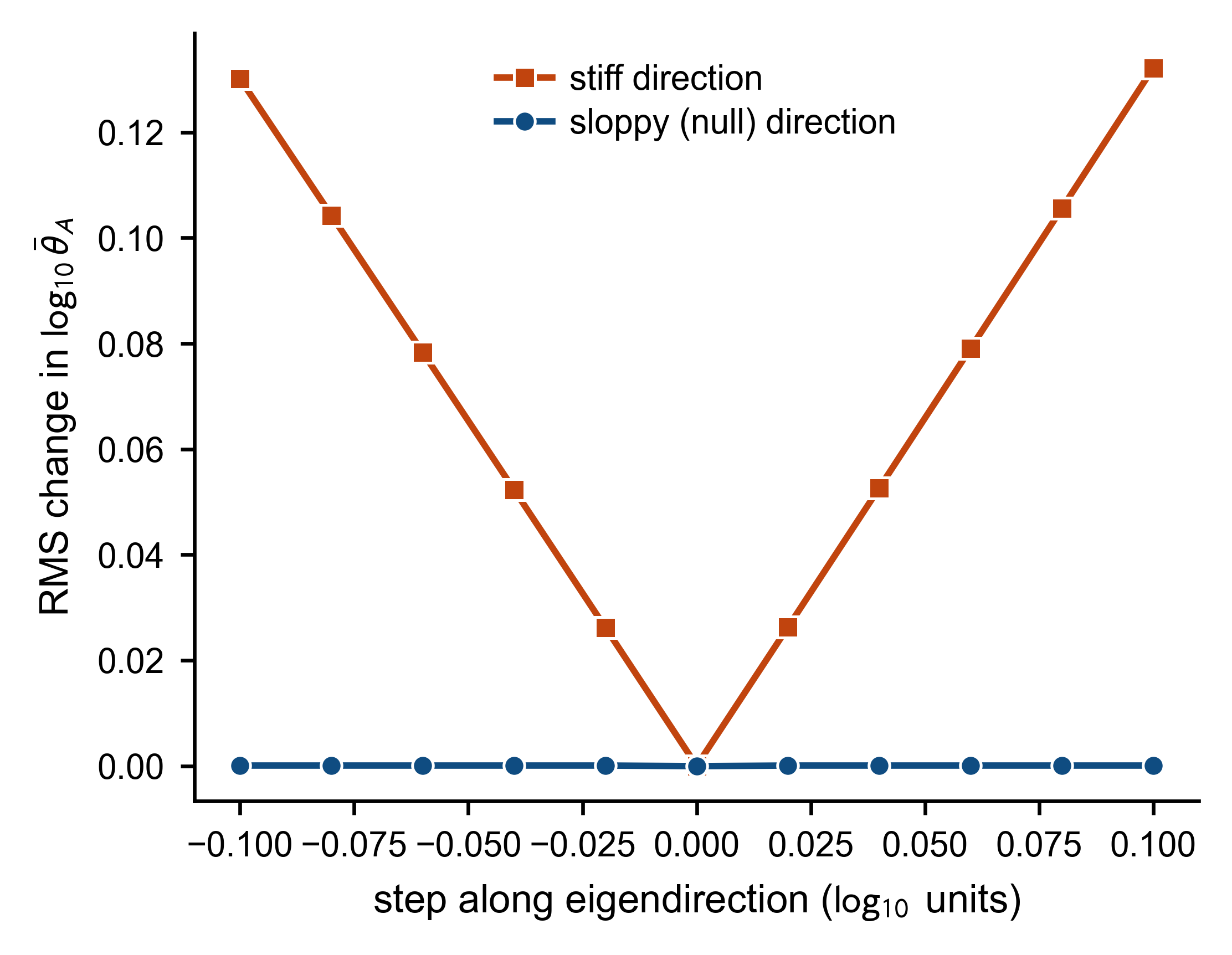}
\caption{Deepened structural analysis. \textbf{(a)}~Eigenvalue spectrum of the
closure-free Fisher matrix---$\lambda_3$ is an exact structural zero at the
double-precision floor ($|\lambda_3|/\lambda_1\!\approx\!2\times10^{-16}$).
\textbf{(b)}~Singular-value spectrum of the prediction-sensitivity matrix
$\partial f_i/\partial\beta_j$, mirroring the hierarchy. \textbf{(c)}~Sloppy-manifold
visualisation---moving $\pm10\%$ along the null direction leaves the model output
essentially unchanged, while the same step along the stiff direction changes it by
orders of magnitude.}
\label{fig:fisherdeep}
\end{figure}

\subsection{Prior ablation: identification does not depend on prior leakage}
The single-temperature inversion includes a weak $\kdes$ prior whose centre is set
at the truth by default, raising the concern that the prior (and a same-magnitude
initialisation) might leak the truth. We therefore re-run the otherwise identical
inversion while varying only the centre and presence of this prior (Table~\ref{tab:prior}).
Two facts rule out leakage: (i) with the prior \emph{removed}, $\Eads=0.772$~eV
(within $0.3\%$ of truth), essentially unchanged; and (ii) even with the prior
centre mis-placed by $\pm2$ decades, $\Eads$ stays within $0.762$--$0.788$~eV, a
spread of only $26$~meV across all settings. $\kdes$ itself is mildly pulled by a
mis-placed prior ($0.58$--$1.58$), but after the log compression of
$\Eads=k_BT\ln(\nu/\kdes)$ this is only $\pm1.5\%$ in $\Eads$. The test $\Rlog$ is
$0.997$ for all settings. Hence $\Eads$ is data-driven, not prior-leaked; the prior
acts only as numerical regularisation.

\begin{table}[t]\centering
\caption{$\kdes$ prior ablation ($w_p=0.01$, 8-seed mean$\pm$s.d.).}
\label{tab:prior}
\small
\begin{tabular}{lccc}
\toprule
prior centre $\log_{10}\kdes^{\mathrm{prior}}$ & $\Eads$ (eV) & $\kdes$ (s$^{-1}$) & $\Eads$ error \\
\midrule
$0$ (truth, default) & $0.773\pm0.001$ & $1.04\pm0.03$ & $-0.1\%$ \\
$-1$ & $0.781\pm0.002$ & $0.76\pm0.06$ & $+0.9\%$ \\
$+1$ & $0.766\pm0.001$ & $1.34\pm0.03$ & $-1.0\%$ \\
$-2$ & $0.788\pm0.003$ & $0.58\pm0.06$ & $+1.9\%$ \\
$+2$ & $0.762\pm0.001$ & $1.58\pm0.07$ & $-1.6\%$ \\
\textbf{none} ($w_p=0$) & $\mathbf{0.772\pm0.001}$ & $1.06\pm0.03$ & $\mathbf{-0.3\%}$ \\
\bottomrule
\end{tabular}
\end{table}

\subsection{Leave-one-out cross-validation}
Because the fixed split places near-saturated cases mostly in training, one may ask
whether the 9-point test $R^2$ is favourably biased. We therefore perform
leave-one-out cross-validation (LOOCV) over all 48 cases: each fold holds out one
case as a clean test and trains on the other 47. The overall LOOCV $\Rlog=0.996$ and
$R^2_{\mathrm{raw}}=0.974$, essentially identical to the fixed split ($0.998/0.989$);
the inverted $\Eads=0.776\pm0.003$~eV and $\kdes=0.94\pm0.09$~s$^{-1}$ also agree with
the main results. The $\sim\!10\%$ shift in $\kdes$ relative to the fixed-split
$1.04\pm0.03$~s$^{-1}$ (Section~6.5) lies within the leave-one-out fold-to-fold spread
($\pm0.09$) and reflects the subset composition of the held-out folds, not an
instability; the identifiable quantity $\Eads$ agrees to within $0.4\%$
($0.776$ vs $0.773$~eV), as expected since the weak-constraint $\kdes$ variation is
compressed to $\pm1.5\%$ in $\Eads$ through $\Eads=k_BT\ln(\nu/\kdes)$. The original
split therefore does not embellish the metrics. The
LOOCV $R^2_{\mathrm{raw}}$ and the maximum log error ($0.187$, at the $\vsub=1.2$
high-speed saturated corner) further expose the saturated-corner model-form bias of
Section~6.1, which LOOCV honestly reveals rather than hides. The anchoring necessity
of ``saturated cases in training'' is reconciled with evaluation fairness because
each training fold still contains all other saturated cases. As a further check at
coarser granularity, $5$- and $10$-fold cross-validation give the same picture, with
the inverted $\Eads$ varying by only $\sim1.5$~meV across folds, confirming that the
inversion is stable to the train/test partition.

\subsection{Computational cost}
The 48-case COMSOL solve (production mesh, 16-core CPU) takes about $17{,}117$~s in
total, i.e.\ $\approx357$~s per case. The one-off PCINN training takes
$\approx1.6$~min ($\approx13$~min for 8 seeds); a trained query needs a single
forward pass: $\approx0.34$~ms amortised per case in batch, $\approx7$~ms standalone. The model is
deliberately tiny: the physics branch is a $2\!\to\!64\!\to\!64\!\to\!64\!\to\!64\!\to\!1$
MLP ($12{,}737$ weights and biases) plus two trainable kinetic scalars, $12{,}739$
parameters in total ($\approx12{,}800$ for the three-input multi-temperature pipeline);
a single coverage evaluation costs only $\approx3\times10^4$ FLOPs
($\approx2.5\times10^4$ for the MLP and $\approx4\times10^3$ for the $320$-step
trajectory integration). The millisecond inference time is therefore dominated by the
serial integration loop rather than by arithmetic, and the small parameter count and
FLOPs are a direct consequence of hard-coding the chemistry and compressing the
data-driven freedom to a single scalar. All computations run on a $16$-core CPU (an
RTX~4070 Laptop GPU is available but the serial chemistry integration does not benefit
from it).
Thus, once the dataset and model are ready, predicting a new case is
$\sim5\times10^4$ (standalone) to $\sim1.1\times10^6$ (amortised) times faster than
CFD. Honestly, generating the training set still incurs the CFD cost, so the
surrogate does not eliminate the up-front investment but amortises it: once the
number of required evaluations exceeds the dataset size (e.g.\ thousands of
evaluations for operating-window optimisation), the advantage is realised---scanning
$10^4$ cases with PCINN takes $\approx3.4$~s versus $\approx42$ days of case-by-case
CFD.

\section{Multi-temperature extension and a model-mismatch test}

At a single temperature, $\Eads$ and $\kdes$ are the same identifiable quantity, so
the single-temperature ``$\Eads$'' is converted from an \emph{assumed} $\nu$.
Separating $\nu$ and $\Eads$ intuitively calls for multi-temperature data, with the
slope and intercept of the Arrhenius line $\ln\kdes(T)$ versus $1/T$ giving $\Eads$
and $\nu$. We test this intuition over four temperatures and characterise the
resulting identifiability faithfully.

\subsection{Multi-temperature dataset and free inversion}
We sweep $T\in\{300,320,340,360\}$~K $\times$ $\vsub$ (8) $\times$ $\Ucur\in\{0.5,2,8\}$
($=96$ cases), with $\kdes\approx\{1.0,6.5,33.7,146\}$~s$^{-1}$ spanning two decades
and remaining saturable throughout. (Higher temperatures, e.g.\ $450$~K, would
collapse the coverage to $\sim10^{-5}$ and lose the saturated cases that anchor
$\kdes$.) The wall shear rates are identical across temperatures, confirming
flow--temperature decoupling. The physics-branch input is augmented to
$(v^*,U^*,T^*)$; the chemistry branch trains $\log_{10}\nu$ and $\Eads$ directly
(dropping the $\kdes$ prior), synthesising $\kdes(T)$ per sample.

With $\thetabar$ over 8 seeds, the surrogate reaches test $\Rlog=0.9991\pm0.0003$
with $\Eads$ cross-seed scatter of only $\pm0.002$~eV---highly reproducible. Yet the
point estimate $\Eads=0.711\pm0.002$~eV is systematically $8\%$ below truth, and
$\log_{10}\nu=11.998\pm0.006$ barely leaves its initial value of $12.0$, i.e.\ $\nu$
is not data-driven but parked at the initialisation, while $\kads$ (effective)
drifts to $0.053$ to absorb the amplitude slack (Table~\ref{tab:multiT}).
\textbf{It must be stated plainly that, under multiple temperatures, the free-inversion
point estimates of $\nu$ and $\Eads$ are not trustworthy and should not be reported
as inversion results; the identifiability conclusion must be given via the profile
likelihood interval or an independent prior (Section~7.3).} The $\Eads=0.711$ in
Table~\ref{tab:multiT} is therefore to be read as ``the parked value of the
optimiser in the degeneracy valley (a non-unique solution)'', not as an estimate of
$\Eads$.

\begin{table}[t]\centering
\caption{Multi-temperature free inversion (label $\thetabar$, 8-seed mean$\pm$s.d.).}
\label{tab:multiT}
\small
\begin{tabular}{lcl}
\toprule
quantity & ground truth & PCINN (8 seeds) \\
\midrule
surrogate $\Rlog$ (test) & --- & $0.9991\pm0.0003$ \\
$\Eads$ (eV) & $0.774$ & $0.711\pm0.002$ (\textbf{parked value, non-unique}) \\
$\log_{10}\nu$ & $13$ & $11.998\pm0.006$ (parked at init) \\
$\kads$ (effective) & $0.01$ & $0.053\pm0.005$ \\
\bottomrule
\end{tabular}
\end{table}

\subsection{A degeneracy valley containing the ground truth}
\textbf{Initial-value drift.} Fixing the $\Eads$ and $\kads$ initial values and
placing $\log_{10}\nu$ at $\{11,12,13\}$ (3 seeds each), the converged $\log_{10}\nu$
follows its initial value (regression slope $0.97$), while the re-optimised $\Eads$
takes $0.652/0.711/0.769$, all with test $R^2>0.998$. The three points are collinear,
giving the degeneracy line $\Eads=0.064\log_{10}\nu-0.061$; extrapolating to
$\log_{10}\nu=13$ gives $\Eads=0.775$~eV, so the ground truth $(10^{13},0.774)$ lies
on the line.

\textbf{Profile likelihood.} Freezing $\log_{10}\nu$ on a grid $[11.0,13.5]$ and
re-optimising the rest, the data-misfit curve is a shallow one-sided valley with a
floor near $13.25$ and a curvature of only $2.2\times10^{-4}$/decade$^2$. The
practical identifiable interval, placed on a principled footing by a $\chi^2_1$
likelihood-ratio criterion rather than an ad-hoc misfit multiple (the profile misfit is
a log-space mean-squared error; with the $\sigma=5\%$ label noise the $95\%$ level
corresponds to a misfit rise of $\chi^2_{1,0.95}\,\sigma^2/N$ above the floor), is
$\log_{10}\nu\in[12.85,13.43]$ ($\approx0.6$ decade, containing the truth
$\log_{10}\nu=13$), giving a lower bound $\nu\gtrsim10^{12.85}$. The previously reported
``$2\times$ floor'' band $[12.5,13.5]$ ($\approx1$ decade) in fact corresponds to a
much higher confidence ($\sim99.99\%$) and is thus a conservative over-estimate of the
width; the $\chi^2_1$ interval is the defensible one. \textbf{Conditioning on $\nu=\nu_{\mathrm{true}}$ recovers
$\Eads=0.776\pm0.001$~eV ($0.3\%$ error)}: given $\nu$, the multi-temperature data
recover $\Eads$ almost exactly; the free-inversion $0.711$ is purely the optimiser
parking low in the valley (small gradient on the shallow slope; classic sloppy
behaviour). The two independent experiments yield the same degeneracy line.

\textbf{Theoretical slope.} The valley is not a numerical artefact but an exact
structural consequence of the Arrhenius form. Fixing the desorption rate at a
reference temperature $T_0$, $\kdes(T_0)=\nu\,e^{-\Eads/k_BT_0}$ implies
$\ln\nu=\ln\kdes(T_0)+\Eads/(k_BT_0)$; holding the data-constrained $\kdes(T_0)$
fixed therefore binds $\Eads$ and $\log_{10}\nu$ along a line of slope
\begin{equation}
\frac{dE_{\mathrm{ads}}}{d\log_{10}\nu}=k_B\,T_{\mathrm{eff}}\,\ln 10 .
\end{equation}
With multi-temperature data the Arrhenius constraint weights $1/T$, so the effective
temperature is the harmonic mean of the sampled temperatures,
$T_{\mathrm{eff}}=\langle 1/T\rangle^{-1}=328.5$~K for $\{300,320,340,360\}$~K, giving
a \emph{predicted} slope of $0.0652$~eV/decade---within $0.7\%$ of the observed
$0.0647$. Three consequences follow: (i) the slope is fixed analytically by the
Arrhenius law and the temperature window, independent of the ground truth, the
network, or noise, so the valley is a structural property rather than a numerical
coincidence; (ii) it predicts that the slope is nearly invariant under any model
mismatch that preserves a \emph{single} Arrhenius process, and shifts only when a
\emph{second} thermally activated process is introduced (both confirmed in
Section~7.3); and (iii) breaking the degeneracy requires enlarging $T_{\mathrm{eff}}$
(a wider temperature lever, which loses the saturated anchoring cases) or supplying
an independent $\nu$ prior. The observed identifiable interval ($\chi^2_1$ $95\%$ width
$\approx0.6$ decade) is thus a direct quantitative consequence of the $60$~K temperature window: with
$T_{\mathrm{eff}}=328.5$~K the slope $k_BT_{\mathrm{eff}}\ln10=0.0652$~eV/decade sets how
far along $\log_{10}\nu$ the valley extends before the data-misfit doubles, so the
interval width and the window are two views of the same quantity---the geometry is
therefore usable as a design tool (Section~8.4), telling in advance how much wider a
temperature range would be needed to separate $\nu$ and $\Eads$. A partial remedy that
requires no new data is to re-parameterise the Arrhenius law about a reference
temperature inside the window, writing $\ln\kdes(T)=\ln\kdes(T_{\mathrm{ref}})
-\tfrac{\Eads}{k_B}\!\left(\tfrac1T-\tfrac1{T_{\mathrm{ref}}}\right)$: this
re-centres the fit on the well-constrained combination $\kdes(T_{\mathrm{ref}})$ and
largely decorrelates the estimated slope ($\Eads$) from the intercept ($\ln\nu$),
tightening the conditional interval without widening $T_{\mathrm{eff}}$. It does not
break the structural degeneracy---the flat direction persists---but it improves its
numerical conditioning, and is the natural finite-data companion to the
manifold-boundary approximation method of Transtrum and Qiu, in which such
degenerate directions are removed by boundary (limiting) reparameterisations rather than
by more data.

\textbf{Generality beyond the strict Arrhenius prefactor.} The slope law depends only
on the activated (exponential) factor and the sampled temperature set, not on the form
of the prefactor. For any generalized rate $\kdes(T)=\nu\,g(T)\,e^{-\Eads/k_BT}$ in
which $g(T)$ is a \emph{known} temperature modulation---modified Arrhenius
$g(T)=T^{n}$, Eyring/transition-state $g(T)=k_BT/h$, or any fixed dependence---the
logarithm $\ln\kdes=\ln\nu+\ln g(T)-\Eads/(k_BT)$ places $g$ in a term that is known
and hence does not enter the $\nu$--$\Eads$ trade-off. The degeneracy direction is set
by the $-\Eads/(k_BT)$ term alone, so the slope remains $k_BT_{\mathrm{eff}}\ln10$ with
the \emph{same} harmonic-mean $T_{\mathrm{eff}}$. The law is thus a property of
single-channel activated-rate inversion in general, not of the strict Arrhenius
prefactor or of ALD; it is altered only by a genuinely different activation
structure---a second exponential (Section~7.3) or a temperature-dependent barrier
$\Eads(T)$---which is exactly what makes the slope a diagnostic for such structure.

\begin{figure}[t]\centering
\includegraphics[width=\textwidth]{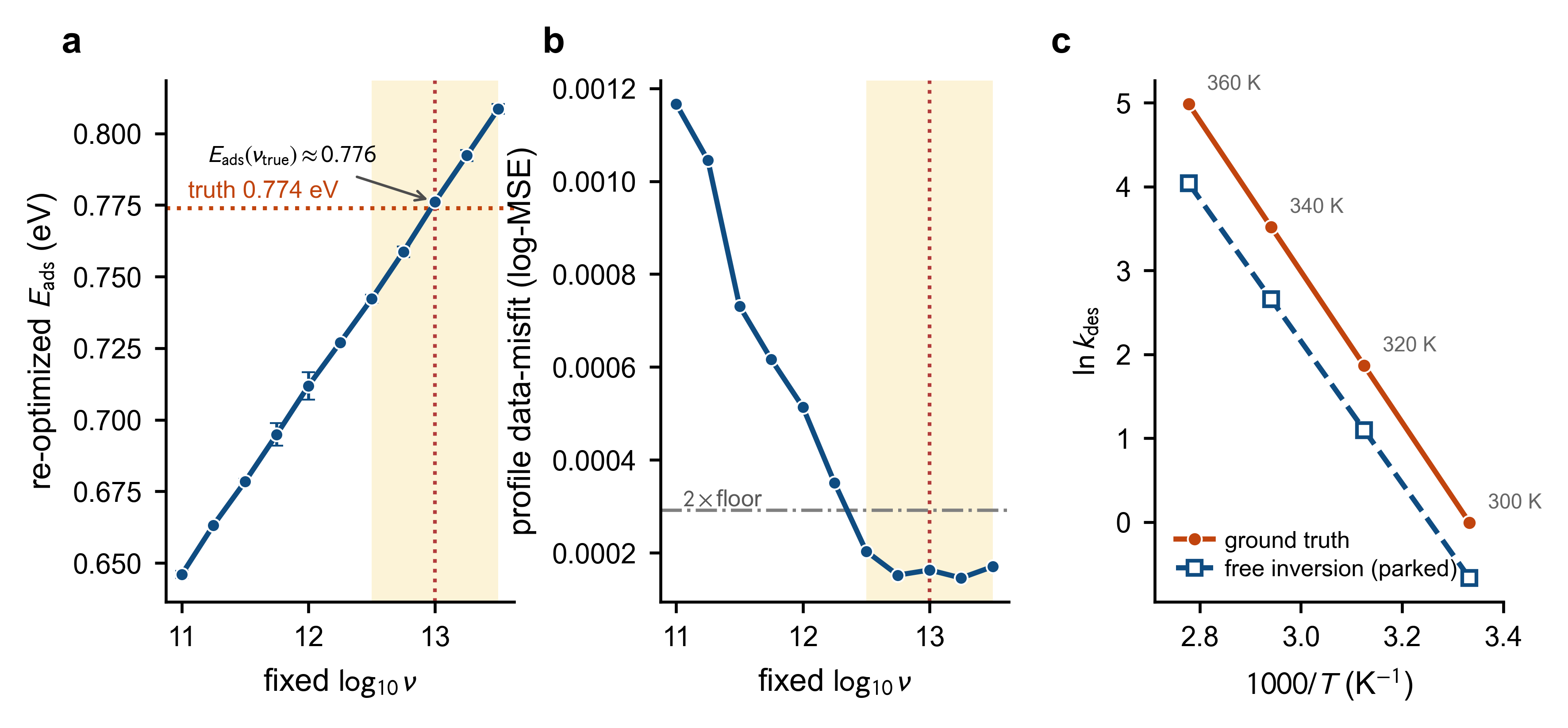}
\caption{Multi-temperature $\nu$--$\Eads$ degeneracy valley.
\textbf{(a)} Re-optimized $\Eads$ versus fixed $\log_{10}\nu$: the solutions slide along
one degeneracy line with the ground truth on it, and conditioning on $\nu_{\mathrm{true}}$
recovers $\Eads\approx0.776$~eV. \textbf{(b)} Profile likelihood in $\log_{10}\nu$---a
shallow one-sided valley; the shaded band is the heuristic ``$2\times$ floor'' region
$\log_{10}\nu\in[12.5,13.5]$, which corresponds to a very high ($\sim99.99\%$)
confidence; the defensible $\chi^2_1$ likelihood-ratio $95\%$ interval is the tighter
$[12.85,13.43]$ ($\approx0.6$ decade, $\nu\gtrsim10^{12.85}$; Section~7.2). \textbf{(c)} Arrhenius plot $\ln\kdes$ vs $1/T$: the
free-inversion line (parked low in the valley) versus the ground truth, which the
$\nu$-conditioned estimate reproduces. The vertical dotted line marks the truth
$\log_{10}\nu=13$; the dotted horizontal line in (a) marks $\Eads=0.774$~eV.}
\label{fig:valley}
\end{figure}

\subsection{Model-mismatch matrix: seven chemistries across three mechanisms}
Since the data are generated and inverted with the same Langmuir--Arrhenius form, we
test whether the conclusions persist when the generating chemistry differs from the
inversion assumption. Rather than a single mismatch, we build a matrix of \emph{seven}
generating chemistries spanning \emph{three orthogonal mechanism families}---desorption-side
coverage dependence (Temkin), adsorption-side non-linearity (Freundlich), and site
heterogeneity (dual-site)---each generated by COMSOL with the \emph{same} underlying
Arrhenius truth and inverted with the \emph{unchanged} standard single-site Langmuir
PCINN (profile likelihood, best-seed). We first detail the canonical desorption-side
case (Temkin $\beta=2$) and then summarise the full matrix (Table~\ref{tab:matrix}).
The Temkin desorption reads
\begin{equation}
J_{\mathrm{net}}^{\mathrm{gen}}=\kads\cwall(1-\theta)
-k_{\mathrm{des},0}\,e^{\beta\theta}\,\Gamma_s\,\theta,
\qquad k_{\mathrm{des},0}=\nu e^{-\Eads/k_BT},\ \beta=2.0,
\end{equation}
where $k_{\mathrm{des},0}$ keeps the same Arrhenius truth and the mismatch enters only
through $e^{\beta\theta}$ (motivated by the fact that the Langmuir assumption of
coverage-independent heat of adsorption generally fails; the Temkin isotherm is the
most common correction). With $\beta=2$, the saturated-corner coverage drops by
$36\%$--$60\%$ relative to Langmuir. We re-run the 96-case scan with this mismatched
chemistry (all converged) and invert with the \emph{unchanged} standard Langmuir
PCINN. Three findings emerge (Table~\ref{tab:mismatch}).

First, the surrogate is robust: $R^2_{\mathrm{raw}}$ drops only from $0.992$ to
$0.987$---the Langmuir form still absorbs most of the Temkin data. Second, the
free-inversion $\Eads$ barely moves ($0.711\to0.705$, only $6$~meV) despite the
strong mismatch, which strongly corroborates that the free-inversion point estimate
is dominated by the parking position in the valley and is not trustworthy; the
mismatch instead leaves a diagnosable signature in the systematic temperature
structure of the per-temperature $\kdes$ bias (the pivot shifts from $\sim320$~K to
$\sim340$~K and the low-temperature positive bias amplifies, $+14.7\%\to+39.2\%$).
Third, and central to the diagnostic, the \textbf{degeneracy valley persists under mismatch}:
the profile degeneracy slope is $0.0645$ (versus $0.0647$ self-consistent), the
conditional inversion gives $\Eads=0.768$~eV (versus $0.776$, only $0.7\%$
degradation), and the valley merely flattens further (curvature
$1.5\times10^{-4}$, interval widening to $2.5$ decades) so that $\nu$ becomes harder
to pin down, without breaking the identifiability geometry.

\begin{table}[t]\centering
\caption{Self-consistent (Langmuir) versus mismatched (Temkin) profile/degeneracy.}
\label{tab:mismatch}
\small
\begin{tabular}{lccl}
\toprule
quantity & Langmuir & Temkin & meaning \\
\midrule
degeneracy slope (eV/decade) & $0.0647$ & $0.0645$ & nearly identical \\
conditional $\Eads$ at $\nu_{\mathrm{true}}$ (eV) & $0.776$ & $0.768$ & only $0.7\%$ degradation \\
valley floor curvature (/decade$^2$) & $2.2\times10^{-4}$ & $1.5\times10^{-4}$ & flatter under mismatch \\
identifiable interval ($2\times$floor, decade) & $1.0$ & $2.5$ & comparison only; cf.\ $\chi^2_1$ $0.6$ dec (\S7.2) \\
\bottomrule
\end{tabular}
\end{table}

\begin{table}[t]\centering
\caption{Model-mismatch matrix: seven core generating chemistries across three mechanism
families, plus two additional dual-site variants (the energy-split calibration sweep,
Section~7.3) and one \emph{transport-layer} mismatch (non-Fickian, concentration-dependent
diffusivity)---each inverted with the \emph{unchanged} single-site Langmuir PCINN
(profile likelihood, best-seed). The degeneracy slope is nearly invariant
($0.063$--$0.070$, theory $0.0652$, Section~7.2) across all six \emph{single-Arrhenius}
chemistries \emph{and} under the transport perturbation; among the dual-site
(double-Arrhenius) cases it is shifted above the band only at the intermediate energy
split ($\Delta E=0.083$), while the small and large splits stay within it and are instead
flagged by fit degradation (Section~7.3, Fig.~\ref{fig:calib}).}
\label{tab:matrix}
\small
\begin{tabular}{llccc}
\toprule
generating chemistry & mismatch mechanism & slope (eV/dec) & cond.\ $\Eads$ (eV) & surrogate $R^2$ \\
\midrule
Langmuir (self-consistent) & --- & $0.0647$ & $0.776$ & $0.998$ \\
\textbf{Langmuir, non-Fickian $D(c)$} & \textbf{transport} & $\mathbf{0.0651}$ & $\mathbf{0.778}$ & $0.998$ \\
Temkin $\beta=2$ & desorption & $0.0645$ & $0.768$ & $0.997$ \\
Temkin $\beta=3$ & desorption & $0.063$ & $0.766$ & $0.997$ \\
Temkin $\beta=4$ & desorption & $0.068$ & $0.782$ & $\mathbf{0.69}$ \\
Freundlich $n=0.5$ & adsorption & $0.066$ & $0.795$ & $0.86$--$0.99$ \\
Freundlich $n=1.5$ & adsorption & $0.070$ & $0.781$ & $0.94$--$0.99$ \\
\textbf{dual-site} ($\Delta E{=}0.083$) & \textbf{site heterogeneity} & $\mathbf{0.0755}$ & $\mathbf{0.811}$ & $0.99$ \\
dual-site ($\Delta E{=}0.04$) & site heterogeneity & $0.0663$ & $0.764$ & $0.998$ \\
dual-site ($\Delta E{=}0.148$) & site heterogeneity & $0.0673$ & $0.736$ & $0.991$ \\
\bottomrule
\end{tabular}
\end{table}

\textbf{The full matrix and a slope-invariance boundary.} Extending beyond the
canonical Temkin case, we add a stronger desorption dependence (Temkin $\beta=3,4$),
an adsorption-side non-linearity (Freundlich, with the adsorption flux
$\propto(1-\theta)^{n}$, $n=0.5,1.5$---the opposite mechanism family), and site
heterogeneity (a dual-site model with two \emph{independent} Arrhenius desorption
channels, $\Eads^{(1)}=0.732$ and $\Eads^{(2)}=0.815$~eV at equal site density).
Table~\ref{tab:matrix} reveals a sharp structural distinction. All six chemistries
that preserve a \emph{single} Arrhenius process---across both the desorption (Temkin)
and adsorption (Freundlich) families---yield a degeneracy slope clustered in
$0.063$--$0.070$ (theory $0.0652$, Section~7.2) and a conditional $\Eads$ within $3\%$
of the truth: the slope is invariant not only to the mismatch strength but to the
mechanism family. The case that moves the slope is the dual-site model,
the only one that superposes a \emph{second} Arrhenius process: at an intermediate
energy split ($\Delta E=0.083$) its slope rises to
$0.0755$ and its conditional $\Eads$ to $0.811$~eV (close to the slow, sticky site
$0.815$ rather than the two-site mean $0.774$), exactly as the $T_{\mathrm{eff}}$
theory of Section~7.2 predicts (an apparent $T_{\mathrm{eff}}\approx380$~K, beyond the
data window). This both confirms the theory and \emph{delineates its boundary}: the
degeneracy slope is invariant under any single-Arrhenius mismatch and shifts when
a second thermally activated process is introduced---though, as the energy-split sweep
below shows, the magnitude of the shift is itself $\Delta E$-dependent, peaking when the
two channels are comparably weighted over the temperature window.

\begin{figure}[t]\centering
\includegraphics[width=0.95\textwidth]{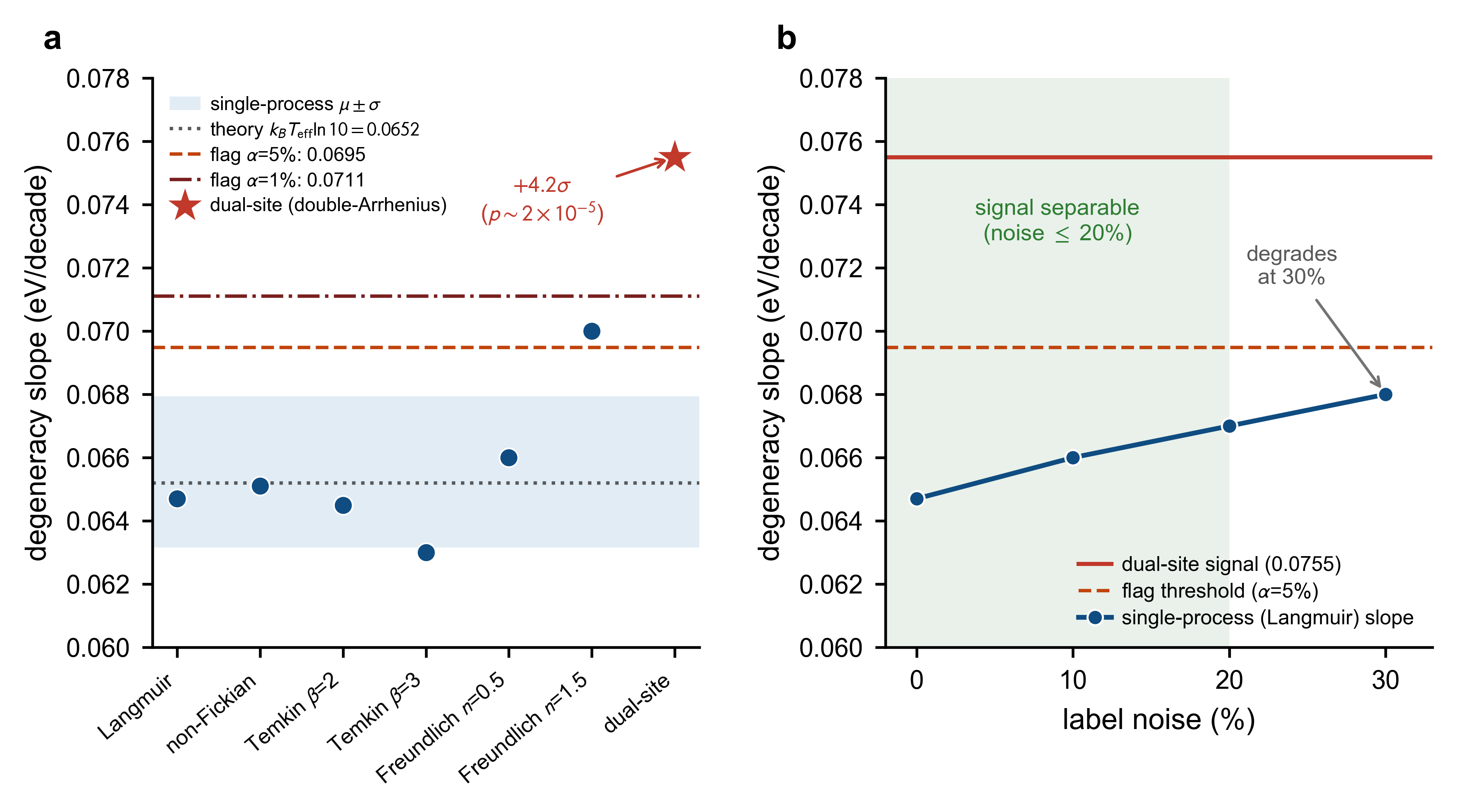}
\caption{Statistical validation of the slope diagnostic. (a) The six single-Arrhenius
generating chemistries (plus the non-Fickian transport mismatch) cluster at
$\mu=0.0655\pm0.0024$~eV/decade; the dual-site (double-Arrhenius) signal at $0.0755$ lies
$+4.2\sigma$ above the cluster (false-positive probability $\sim1.5\times10^{-5}$), well
clear of the $\alpha=5\%$ ($\mu+1.64\sigma$) and $\alpha=1\%$ ($\mu+2.33\sigma$) flag
thresholds. (b) Noise robustness: the single-process slope stays below the $5\%$
threshold up to $20\%$ label noise while the dual-site signal remains separable,
degrading only at $30\%$.}
\label{fig:diagnostic}
\end{figure}

\textbf{An operational threshold with a controlled false-positive rate.} This turns the
slope into a concrete, statistically calibrated test rather than a qualitative
observation. The six single-Arrhenius chemistries (plus the non-Fickian transport
mismatch), spanning both the desorption and adsorption mechanism families, form a
tight reference cluster: mean slope $\mu=0.0655$~eV/decade with scatter
$\sigma=0.0024$ (range $0.063$--$0.070$; the collapsed-fit Temkin $\beta=4$ row is
excluded as non-quantitative). The dual-site signal at $0.0755$ lies
$+4.2\sigma$ above this cluster mean---a separation corresponding to a single-process
false-positive probability of $\sim1.5\times10^{-5}$, not a marginal $\sim2\times$ gap
(Fig.~\ref{fig:diagnostic}a). We therefore replace the earlier ``$\sim10\%$'' rule with
a threshold that controls the false-positive rate directly: flag a second thermally
activated process when the empirical slope exceeds $\mu+1.64\sigma=0.0695$~eV/decade
($5\%$ false-positive rate) or, more conservatively, $\mu+2.33\sigma=0.0711$~eV/decade
($1\%$)---consistent with, and now giving a statistical basis to, the previously stated
$\gtrsim0.072$~eV/decade. The diagnostic is also robust to noise: adding
$10\%$/$20\%$/$30\%$ label noise (Section~7.4) leaves the single-process slope at
$0.066$/$0.067$/$0.068$, below the $5\%$ threshold up to $20\%$ noise while the dual-site
signal stays clearly above it (Fig.~\ref{fig:diagnostic}b), so the test retains its
discriminating power over a realistic noise budget. Because the warning is computed from
the profile geometry rather than the fit residual, it fires precisely in the insidious
case where the surrogate $R^2$ remains excellent and no other symptom is visible.

\begin{figure}[t]\centering
\includegraphics[width=0.98\textwidth]{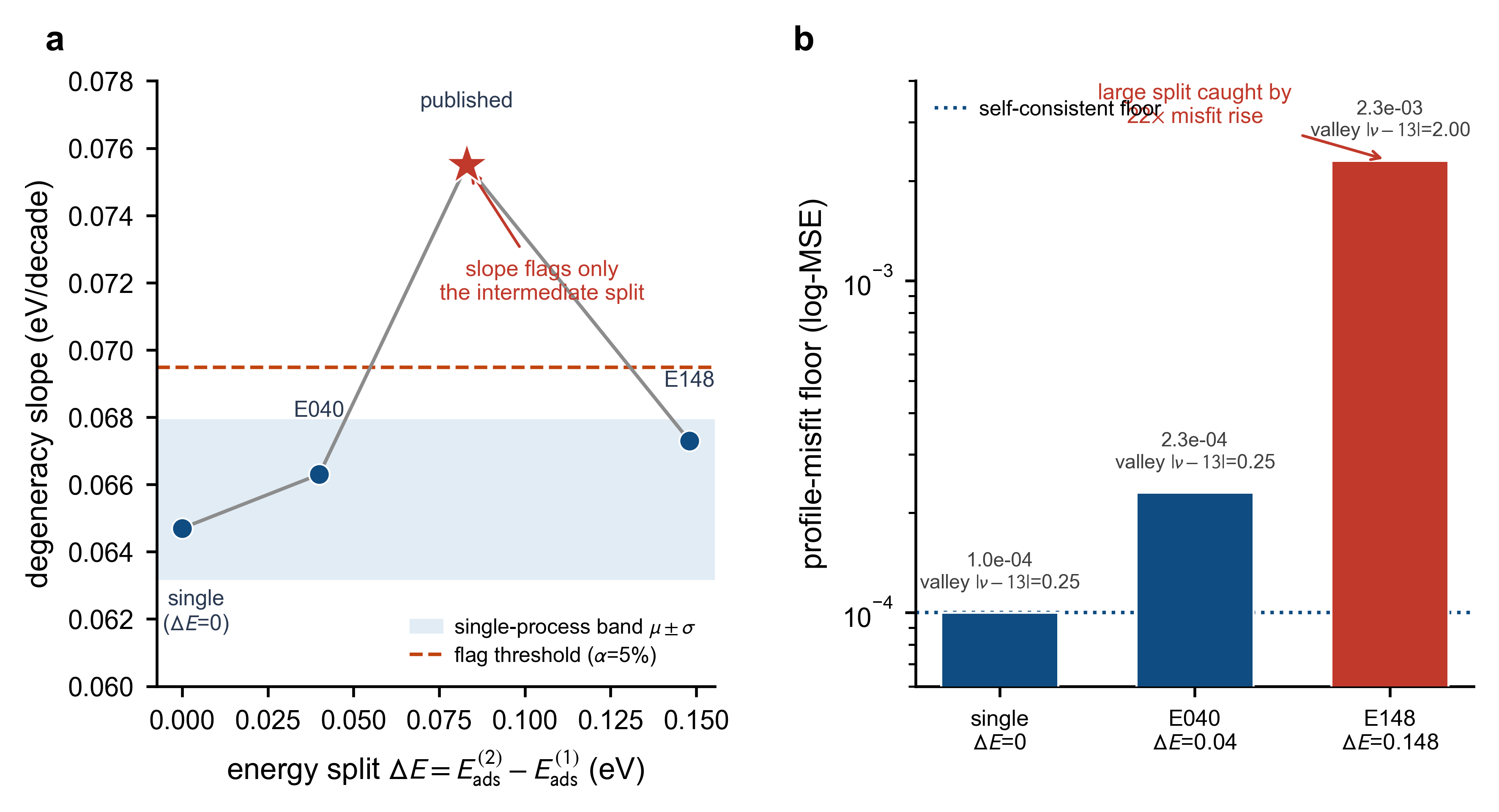}
\caption{Energy-split calibration and multi-signal detection. (a) Degeneracy slope
versus the dual-site energy split $\Delta E$: the excursion above the single-process band
($\mu\pm\sigma$) peaks at the intermediate split ($\Delta E=0.083$, slope-flagged) and
returns into the band at the small ($0.04$) and large ($0.148$) splits---a detection
window predicted by the $T_{\mathrm{eff}}$ theory. (b) Complementary signal: the
profile-misfit floor. The large split ($\Delta E=0.148$), which the slope does not flag,
raises the misfit floor $\sim\!22\times$ over the self-consistent value and pushes the
$\nu$-valley to the grid edge (two decades from truth), while the near-degenerate small
split ($\Delta E=0.04$) is benign on every signal. The slope and the fit geometry are
thus complementary, covering intermediate and large splits respectively.}
\label{fig:calib}
\end{figure}

\textbf{An energy-split calibration sweep, and a multi-signal detection panel.} To test
the slope beyond the single dual-Arrhenius realisation, we generated two further
equal-density two-site datasets bracketing it, with energy splits
$\Delta E=E_{\rm ads}^{(2)}-E_{\rm ads}^{(1)}=0.04$ and $0.148$~eV (mean fixed at the
$0.774$~eV truth), and inverted each with the unchanged single-site PCINN. The result
is a \emph{detection window} (Fig.~\ref{fig:calib}a): the slope excursion above the
single-process band is maximal at the intermediate split ($\Delta E=0.083$,
$0.0755$, $+4.2\sigma$) and returns \emph{into} the band at both the small split
($0.04\!\to\!0.0663$) and the large split ($0.148\!\to\!0.0673$). This is exactly what
the $T_{\mathrm{eff}}$ theory predicts: the apparent $T_{\mathrm{eff}}$ (and hence the
slope) is raised only when the two channels are \emph{comparably weighted} over the
sampled window; near degeneracy ($\Delta E\!\to\!0$) there is little heterogeneity to
detect, and at large $\Delta E$ one channel dominates the observable and the apparent
behaviour reverts toward single-Arrhenius. The slope is therefore a \emph{specific but
not universally sensitive} flag---a departure implies heterogeneity, but the absence of
one does not exclude it.

The two splits the slope does not flag are nonetheless not missed by the inversion as a
whole, because mismatch is diagnosed by a \emph{panel} of signals, not the slope alone
(Fig.~\ref{fig:calib}b). The small split ($\Delta E=0.04$) is near-degenerate and
genuinely benign: a single-site fit recovers the conditional $\Eads$ to within $1.3\%$
($0.764$ vs $0.774$), with a profile-misfit floor ($2.3\times10^{-4}$) and valley
location ($\log_{10}\nu\!=\!13.25$) indistinguishable from the self-consistent case---so
no signal fires, correctly. The large split ($\Delta E=0.148$), which the slope misses,
is caught instead by the \emph{fit geometry}: its profile-misfit floor is $\sim\!22\times$
the self-consistent value ($2.3\times10^{-3}$ vs $1.0\times10^{-4}$) and its valley floor
is pushed to the edge of the $\nu$ grid ($\log_{10}\nu\!=\!11.0$, two decades from the
truth), both strong departures absent in the self-consistent and small-split cases. Thus
the slope excursion covers the intermediate-split regime where the fit stays excellent
(the insidious case), while fit degradation and valley-floor displacement cover the
large-split regime; together the panel flags site heterogeneity across the $\Delta E$
range wherever it produces a non-negligible bias, and stays appropriately silent where a
single-site description is adequate.

The matrix also exposes two complementary mismatch signatures. Strong
\emph{single-process} mismatch (Temkin $\beta=4$, Freundlich $n=0.5$) is self-reporting
through a collapse of the surrogate fit ($\Rlog$ falling from $0.997$ to
$0.69$--$0.86$): the inversion warns of its own invalidity. For these collapsed-fit
rows the tabulated slope and conditional $\Eads$ (e.g.\ $0.068$ and $0.782$~eV for
Temkin $\beta=4$ at $\Rlog=0.69$) should be read only qualitatively---they enter
Table~\ref{tab:matrix} to corroborate that a strong single-process mismatch self-reports
through fit collapse, not as reliable slope estimates on par with the high-$R^2$ rows.
Site heterogeneity, by
contrast, is the most insidious---it retains an excellent fit ($R^2>0.99$, no warning)
yet biases the conditional $\Eads$ by $+4.8\%$ and steepens the slope by $16\%$.
Together these bracket when an inversion result can be trusted: a high $R^2$ is
necessary but not sufficient, and the residual/slope structure (not the fit quality)
is the reliable mismatch diagnostic.

\textbf{A transport-layer mismatch.} The chemistry matrix above perturbs only the
surface-kinetic closure; the transport model (Fickian dilute-species diffusion) is
shared between the data-generating CFD and the PCINN forward map. To probe this
remaining shared assumption, we regenerate the $96$-case multi-temperature dataset
with a \emph{non-Fickian}, concentration-dependent diffusivity
$D_{\mathrm{eff}}(c)=D_{\mathrm{TMA}}(1-\tfrac12 c/C_0)$---a structural transport-model
error that alters the operating-condition$\to$near-wall-concentration mapping the
physics branch must learn---and re-run the \emph{unchanged} Langmuir inversion. Under
this transport perturbation the profile degeneracy slope is $0.0651$~eV/decade,
essentially unchanged from the $0.0647$ of the self-consistent case at the same full
precision (both round to the $0.065$ quoted in Tables~\ref{tab:mismatch}--\ref{tab:matrix}
and agree with the theoretical $0.0652$), and the conditional $\Eads$ at the true $\nu$ is $0.778$~eV
(versus $0.776$), with an identical practical interval ($\log_{10}\nu\in[12.75,13.50]$)
and surrogate quality ($\Rlog\ge0.998$); of the $96$ cases, $87$ converged, the nine
failures clustering at the highest-shear corner ($U_{\mathrm{curtain}}=8$), where the
non-Fickian term most stiffens the near-wall layer. The learned closure thus
\emph{absorbs} the transport error: because the physics branch fits the
operating-condition$\to c_{\mathrm{wall}}$ map directly from data, it simply learns the
new (non-Fickian) mapping, and the kinetic inversion---which reads $\Eads$ from the
\emph{temperature slope} of $\thetabar$---is left unbiased. This outcome is structural
rather than fortuitous, and it fixes the precise scope of the test: the perturbation is
\emph{temperature-independent}, so it rescales the near-wall concentration amplitude
without touching its Arrhenius temperature scaling. The experiment should therefore be
read as a \emph{sufficiency proof for amplitude-type (temperature-independent) transport
mismatch}---it establishes that the physics-branch closure absorbs any transport error
that leaves the temperature dependence intact, and hence cannot, by construction, move
the degeneracy slope. It is deliberately not a test of the harder case: a
\emph{temperature-dependent} transport error---whose physical instances include a
temperature-dependent gas diffusivity $D(T)$ or viscosity $\mu(T)$, both realistic in a
heated SALD gap---would introduce an additional
temperature channel and, exactly like a second thermally activated process
(Section~7.2), \emph{would} be expected to shift the slope---the same falsifiable
signature the diagnostic is designed to catch. That case delineates the boundary of the
present test and is left to future work alongside validation on real data. Within its
stated scope the result is nonetheless a genuine extension beyond the chemistry matrix:
it demonstrates that the identifiability geometry and slope diagnostic survive a
mismatch in the transport layer, not only the chemistry, without substituting for
validation against experimental data.

The structural conclusions of the multi-temperature identifiability analysis---the
existence of the degeneracy valley, its slope, and the ``condition on $\nu$ then
$\Eads$ is recoverable'' reporting convention---thus hold under moderate model
mismatch. The effect of mismatch is a further flattening of the valley and a mild
($0.7\%$) degradation of the conditional $\Eads$, not a breakdown of the
identifiability geometry.

\begin{figure}[t]\centering
\includegraphics[width=0.6\textwidth]{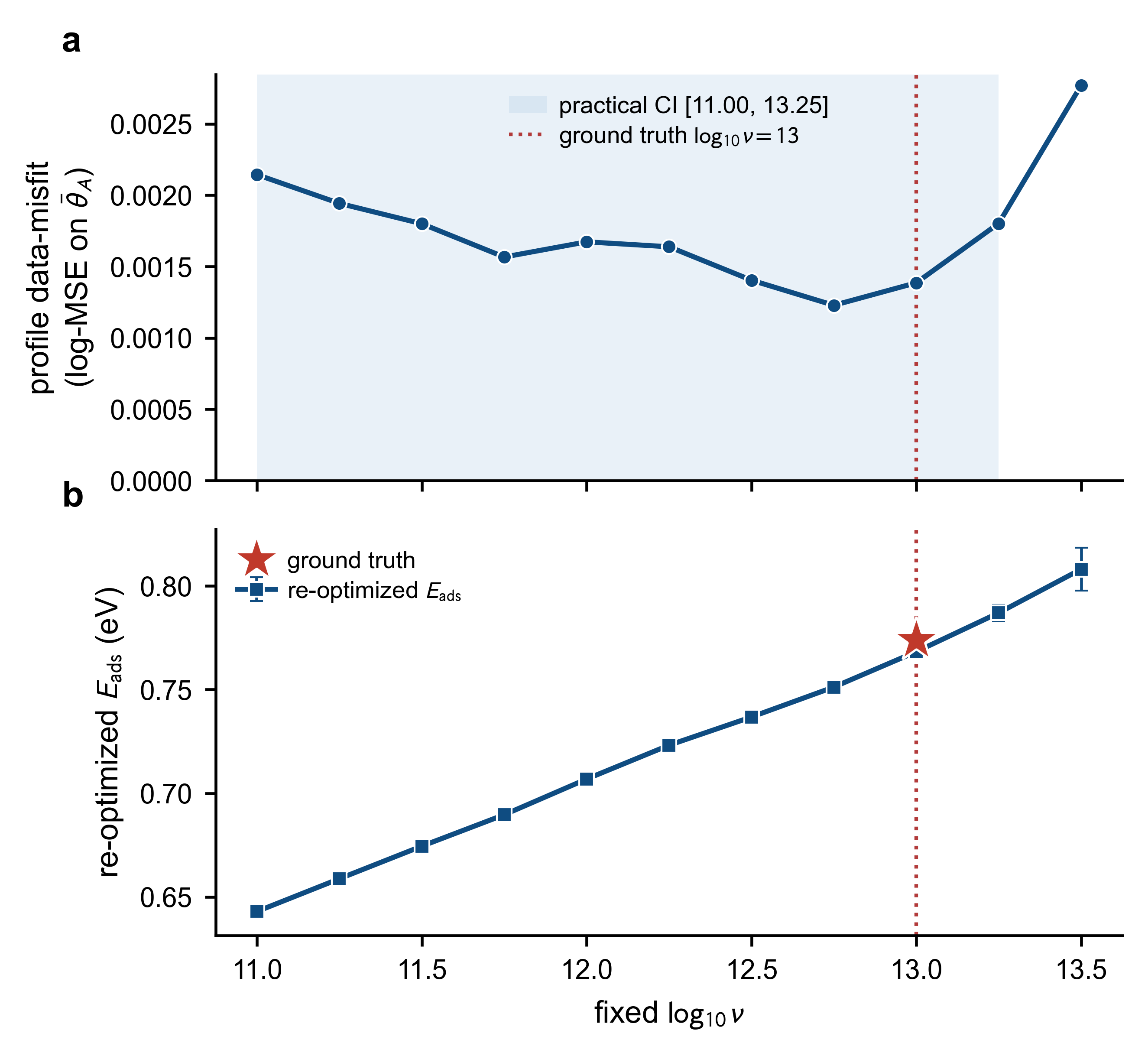}
\caption{Degeneracy valley under model mismatch (profile likelihood in
$\log_{10}\nu$). Across the seven-chemistry matrix the valley and its slope persist;
the conditional $\Eads$ at $\nu_{\mathrm{true}}$ degrades by at most a few percent for
single-Arrhenius mismatches, and only the dual-site (double-Arrhenius) case shifts the
slope (Table~\ref{tab:matrix}).}
\label{fig:mismatchprofile}
\end{figure}

\subsection{Robustness: noise, bootstrap resampling, and coarse-grid sampling}
Three further perturbations confirm that the degeneracy geometry is not an artefact of
the specific regular grid or of the clean labels. (i)~\emph{Label noise.} Adding
$10\%/20\%/30\%$ Gaussian relative noise to the coverage labels leaves the slope at
$0.066/0.067/0.068$ up to $20\%$; only at $30\%$ does the inversion degrade, bounding
the noise tolerance. (ii)~\emph{Bootstrap resampling.} Drawing $30$ training cases at
random, repeated $20$ times, gives a conditional $\Eads=0.785\pm0.012$~eV
(median $0.777$), so the result is insensitive to the training-subset composition
rather than to one favourable split. (iii)~\emph{Coarse-grid, irregular sampling.} To
rule out that the result depends on the fine regular grid---and to address the concern
that a dense regular scan resembles interpolation---we coarsen the CFD mesh (maximum
element $80~\mu$m, tolerance $10^{-3}$, $\approx74$~s per case, about $5\times$ cheaper)
and draw $150$ Latin-hypercube samples over $(\vsub,\Ucur,T)$. The degeneracy slope is
$0.063$ (versus $0.065$ on the fine regular grid), the conditional $\Eads$ extrapolates
to $0.777$~eV at $\nu_{\mathrm{true}}$, and the valley remains flat in $\nu$. The
identifiability geometry is thus independent of mesh resolution and of sampling
regularity. We deliberately did \emph{not} augment the data with a machine-learning
generator: synthesising data from a model trained on the same cases would form a
circular validation loop and add no independent information, whereas bootstrap and
coarse-grid CFD provide genuine robustness checks without that methodological flaw.

\section{Conclusions and outlook}

\subsection{Conclusions}
We established a saturable-window SALD model and a physics--chemistry-informed
neural network that, under sparse data, achieves a high-accuracy coverage surrogate
(test $\Rlog=0.998$ with 30 cases) and a robust adsorption-energy inversion, while
transparently delineating the identifiability and extrapolation boundaries. The
methodological core is that \textbf{the precision boundary of PCINN inversion is set
by the degeneracy structure of parameter space, not by fitting power}: at a single
temperature $\kads$ is not separately identifiable (only $\kads\cwall$ is); under
multiple temperatures $\nu$ and $\Eads$ are bound along a weakly identifiable
degeneracy valley (slope $\approx0.065$~eV/decade, truth on the line), with
$\Eads$ recoverable to $0.3\%$ conditional on $\nu$. We state as a main result, not a
caveat, that \textbf{multiple temperatures do not, in this setting, separate $\nu$ and
$\Eads$}: they only compress the pair into a weakly identifiable interval ($\chi^2_1$ $95\%$ width $\approx0.6$ decade),
and genuine separation requires either an independent $\nu$ prior or spatially
resolved coverage supervision---a deliberately negative finding that we consider as
informative as a positive one. A seven-chemistry,
three-mechanism mismatch matrix, together with three independent lines of evidence for
the slope---its analytic derivation $k_BT_{\mathrm{eff}}\ln10$ (Section~7.2), bootstrap
resampling, and coarse-grid regeneration (Sections~7.3--7.4)---show these
conclusions persist, the slope shifting only when a second Arrhenius process is added.
A prior
ablation (mis-placed or removed) and LOOCV confirm that the identification does not
depend on prior leakage or on the split.

\subsection{Limitations}
($i$)~$\kads$ is an effective value under a weak prior; only $\kads\cwall$ and
$\Eads$ are identifiable at a single temperature. ($ii$)~The near-wall concentration
supervision uses a surface/line average rather than a single-point value (the latter
does not converge in the strong-adsorption regime). ($iii$)~Multiple temperatures do
not separately calibrate $\nu$ and $\Eads$ but narrow them to a sub-decade ($\approx0.6$-decade, $\chi^2_1$ $95\%$)
weakly identifiable valley; the absolute calibration of $\nu$ still requires an
independent prior or spatially resolved profile supervision. ($iv$)~The two-segment
integration assumes $\cwall=0$ downstream of the A zone; under a weak curtain
(low $\Ucur$), precursor cross-talk downstream may slightly under-estimate the
coverage, partly co-sourced with the saturated-corner under-estimation; likewise the
monotonicity prior assumes a monotone near-wall concentration in $\Ucur$ and would
bias the result if real flow features (e.g.\ recirculation-induced re-adsorption)
were non-monotone. ($v$)~The model is single-species (S1) and 2-D steady; the data
are synthetic ``measurements'' without real process data or experimental uncertainty.
Regarding the choice of synthetic ground truth ($\Eads=0.774$~eV, $\nu=10^{13}$~s$^{-1}$):
these lie within the DFT range and the transition-state-theory magnitude
respectively, and the identifiability-related conclusions do not depend on the
specific values---changing the truth only translates the valley in parameter space
without altering its existence or orientation (the mismatch test confirms the slope
is nearly unchanged even under a different generating chemistry).

\subsection{Outlook}
Pinning down $\nu$ absolutely and breaking the $\kads\cwall$ degeneracy require new
information sources---spatially resolved coverage/concentration-profile supervision,
or an independent physical prior for $\nu$---rather than widening the temperature
window (which would lose the saturated anchoring cases). A two-species (S2) extension
would yield the true bidirectional cross-talk and operating window. Conceptually, the
degeneracy slope law is expected to carry over to such coupled half-reactions with an
important qualification: as long as each half-reaction desorbs through a
\emph{single} Arrhenius channel, its own $\nu$--$\Eads$ valley retains the slope
$k_BT_{\mathrm{eff}}\ln10$ set by the shared temperature window, and the inversion factorises
per species. Coupling between successive adsorption--desorption cycles (e.g.\ a
coverage-dependent sticking probability of one precursor on sites conditioned by the
other) enters as an effective coverage dependence of the rate---formally the same
structure as the Temkin/Freundlich mismatches of Section~7.3, which leave the slope
invariant---so it would perturb the valley floor and the conditional $\Eads$ without
moving the slope, \emph{unless} the coupling introduces a second thermally activated
timescale, in which case the slope shifts exactly as in the dual-site case (an apparent
$T_{\mathrm{eff}}$ departure). The single-species diagnostic therefore extends to the
two-species setting as a per-channel test, and a slope anomaly would flag genuinely
coupled (double-Arrhenius) inter-cycle chemistry. The present study is deliberately scoped as a \emph{concept verification} on a compact,
self-consistent benchmark (30 training cases; leave-one-out $R^2_{\mathrm{raw}}=0.974$
controls over-fitting), so that the identifiability analysis is not confounded by data
volume. Scaling to substantially larger datasets and to \emph{field-level}
generalisation---predicting full coverage/concentration fields rather than the
trajectory-averaged $\thetabar$, e.g.\ via a Fourier neural operator (FNO) surrogate---is
left to a dedicated follow-up, of which this paper is the identifiability-focused first
part. Applying the
pipeline to real SALD coverage/thickness data, under experimental noise and model
mismatch, is the key next step towards engineering use.

\subsection{Implications for the methodological community}
The identifiability findings have significance beyond SALD. The $\kads\cwall$
degeneracy (single-temperature) and the $\nu$--$\Eads$ degeneracy valley
(multi-temperature) are instances of a widely known structure: multi-parameter
nonlinear models often exhibit very weak sensitivity along a few ``sloppy''
directions, rendering parameter combinations along them practically non-identifiable
even when the overall fit is excellent.

\emph{Relation to prior work, and our increment.} The sloppy-model geometry of
Transtrum, Machta and Sethna and the profile-likelihood framework of Raue, Kreutz
and co-workers already established the vocabulary of degeneracy valleys, profile
intervals and conditional identifiability that we use here; our analysis lives
largely within that framework rather than replacing it. Rather than claiming priority
for applying identifiability analysis \emph{per se}---profile likelihood and Fisher
analysis are long established on mechanistic ODE models in systems biology---our
increment is a specific \emph{combination} that this literature does not address.
First, we use these diagnostics to locate the \emph{action boundary of a learned
transport closure} feeding a hard-coded kinetic integrator: the negative control of
Section~6.4 shows that the embedded physics confers extrapolation power only along the
structurally known (residence-time) axis and none along the data-driven transport
axis---a delineation specific to hybrid (known-physics-plus-learned-closure) models and
absent from the generic sloppy-model picture. Second, we turn the degeneracy slope from
a descriptive feature into a \emph{predictive, falsifiable diagnostic} via the analytic
law $k_BT_{\mathrm{eff}}\ln10$ and its invariance boundary. We are explicit that the
derivation of the slope itself is elementary---it is a one-line differentiation of the
Arrhenius relation, not a new result---so the contribution is not the algebra but its
conversion into a \emph{statistically validated} test: a threshold with a controlled
false-positive rate, a $+4.2\sigma$ signal-to-cluster separation, and a characterised
noise budget (Section~7.3, Fig.~\ref{fig:diagnostic}), together with the delineation of
where the hybrid closure's embedded physics does and does not help. It is the pairing of
these two---an action-boundary localisation for a learned closure and a physics-derived,
\emph{empirically calibrated} degeneracy test---rather than the underlying algebra, that
constitutes the novelty. We give three cautions for PINN/scientific-ML
inversion. First, \emph{a high goodness of fit does not imply parameter
identifiability}: here the surrogate $\Rlog$ reaches $0.999$ yet some parameters
($\kads$ at a single temperature, $\nu$ at multiple temperatures) are fundamentally
non-identifiable. Second, \emph{a free-inversion point estimate may merely be the
optimiser's parking position in a degeneracy valley}, drifting with the
initialisation and statistically meaningless; the responsible practice is to
characterise the identifiable directions via profile likelihood/Fisher analysis and
report parameters as intervals or conditional on a stated prior. Third, \emph{model
mismatch need not manifest as a drop in goodness of fit but may hide in the residual
structure} (here, the systematic temperature dependence of the per-temperature
$\kdes$ bias); diagnosing mismatch should examine the residual structure, not only
the overall $R^2$.

\textbf{A practitioner checklist.} These cautions translate into a concrete procedure
for anyone inverting kinetics from coverage (or analogous) data with a hybrid model.
(i) Before trusting any point estimate, run a profile-likelihood scan over each target
parameter and report the interval where the misfit stays within a small multiple of
its floor, not a single number. (ii) If a parameter is only weakly identifiable
(a shallow valley), do not report it as a calibrated value; instead report it
\emph{conditional on} an explicitly stated independent prior (e.g.\ a DFT/TST estimate
of $\nu$), and propagate that prior's uncertainty through the conditional estimate.
(iii) Use the geometry as a design tool: the slope law $k_BT_\mathrm{eff}\ln10$ tells
how much a wider temperature window (larger $T_\mathrm{eff}$) would tighten the valley,
quantifying the experimental effort needed to separate $\nu$ and $\Eads$. (iv) Treat a
high $R^2$ as necessary but not sufficient, and screen for model mismatch by inspecting
the \emph{structure} of the residuals---here, the temperature dependence of the
per-temperature $\kdes$ bias and the value of the degeneracy slope (a slope departing
from $k_BT_\mathrm{eff}\ln10$ flags an unmodelled second activated process, e.g.\ site
heterogeneity), rather than the aggregate fit metric alone.
(v)~Do not augment a sparse training set with samples drawn from the same model
class being validated: this manufactures a circular validation loop in which the
generator injects no independent information. We deliberately avoided ML-based
data augmentation for this reason, using instead bootstrap resampling and a
coarse-grid Latin-hypercube redraw (Section~7.4) to test robustness without
compromising the methodology.

\textbf{What transfers to the computational-physics reader.} Stripped of the ALD
specifics, this paper contributes a reusable workflow for \emph{any} inverse problem
that couples a known governing law to a learned closure (a hybrid neural-ODE / UDE or
gray-box model). The workflow is: (1) invert with the known-physics-plus-learned-closure
forward map; (2) map each suspect parameter's degeneracy by profile likelihood (and its
local curvature by the Fisher information); (3) where the governing law admits it,
\emph{derive} the analytic degeneracy direction---here $k_BT_{\mathrm{eff}}\ln10$, a
quantity fixed by the physics and the sampling design alone---and compare it to the
empirical slope; (4) read a match as confirmation that the degeneracy is the expected
structural one (report conditional estimates and use the geometry to design a
degeneracy-breaking measurement), and a departure beyond a stated tolerance as a
falsifiable flag for unmodelled structure, even at excellent fit. Steps (1)--(2) are
generic; the value added here is step (3)---an \emph{analytic}, physics-derived
degeneracy direction that upgrades the usual empirical sloppy-model picture into a
predictive, testable diagnostic. Any activated-rate or otherwise structurally
degenerate inversion---not only surface kinetics---inherits the same construction.

\appendix
\section*{Appendix A. Nomenclature}
\addcontentsline{toc}{section}{Appendix A. Nomenclature}
\small
\begin{longtable}{@{}ll@{}}
\toprule
Symbol & Meaning (unit) \\
\midrule
\endhead
$\theta(x)$, $\thetabar$ & local / substrate-averaged A-zone coverage (--) \\
$\theta_{\mathrm{peak}},\theta_{\mathrm{exit}}$ & peak / exit coverage along the trajectory (--) \\
$\kads$ & adsorption rate constant (m\,s$^{-1}$); truth $0.01$ \\
$\kdes$ & desorption rate constant (s$^{-1}$); truth $1.0$ at $300$~K \\
$\Eads$ & adsorption energy (eV); truth $0.774$ \\
$\nu$ & Arrhenius pre-exponential factor (s$^{-1}$); truth $10^{13}$ \\
$\cwall$ & near-wall precursor concentration (mol\,m$^{-3}$) \\
$C_s^{*}$ & learned effective near-wall concentration scale (--) \\
$C_s^{\mathrm{mid}}$ & COMSOL surface-/line-averaged near-wall conc.\ (mol\,m$^{-3}$) \\
$\Gamma_s$ & saturation site density (mol\,m$^{-2}$); $5\times10^{-6}$ \\
$C_0$ & inlet precursor concentration (mol\,m$^{-3}$); $0.2$ \\
$D$ & gas-phase diffusivity (m$^2$\,s$^{-1}$); $6\times10^{-6}$ \\
$J_{\mathrm{net}}$ & net adsorption flux (mol\,m$^{-2}$\,s$^{-1}$) \\
$T,\,T^{*}$ & temperature (K); normalised $T^{*}=T/300$ \\
$T_{\mathrm{eff}}$ & effective (harmonic-mean) temperature, $\langle 1/T\rangle^{-1}$ (K) \\
$\vsub,\,v^{*}$ & substrate velocity (m\,s$^{-1}$); $v^{*}=\vsub/2.0$ \\
$\Ucur,\,U^{*}$ & curtain velocity (m\,s$^{-1}$); $U^{*}=\Ucur/16$ \\
$k_B$ & Boltzmann constant; $8.617\times10^{-5}$~eV\,K$^{-1}$ \\
$\beta$ & Temkin coverage-dependence exponent (--) \\
$n$ & Freundlich exponent in $(1-\theta)^n$ (--) \\
$\Rlog,\,R^2_{\mathrm{raw}}$ & coefficient of determination in $\log_{10}\theta$ / raw $\theta$ \\
$w_C,w_m,w_p$ & loss weights (concentration, monotonicity, prior) \\
\bottomrule
\end{longtable}
\normalsize

\section*{Appendix B. PCINN forward pass and training (pseudocode)}
\addcontentsline{toc}{section}{Appendix B. PCINN pseudocode}
\small
\begin{verbatim}
# --- Forward: operating condition -> coverage (fully differentiable) ---
# globals: C0, Gamma_s, kB, L_A, L_down, N_A, N_down;  per-case: v_sub, T
def coverage(v_star, U_star, T_star, params):        # params = (log10_kads,
    C_s    = softplus(MLP(v_star, U_star, T_star))   #           log10_nu, E_ads)
    c_wall = C_s * C0
    k_ads  = 10**params.log10_kads
    k_des  = params.nu * exp(-params.E_ads / (kB * T))   # Arrhenius (multi-T)
    theta, integral = 0.0, 0.0
    dx = L_A / N_A                                   # A-zone: c_wall = C_s*C0
    for _ in range(N_A):
        J        = k_ads*c_wall*(1 - theta) - k_des*Gamma_s*theta
        theta   += (dx / (Gamma_s*v_sub)) * J        # forward Euler
        integral += theta * dx
    dx = L_down / N_down                             # downstream: c_wall = 0
    for _ in range(N_down):
        J        = -k_des*Gamma_s*theta
        theta   += (dx / (Gamma_s*v_sub)) * J
        integral += theta * dx
    return integral / (L_A + L_down)                 # -> theta_bar_A

# --- Training ---
for seed in range(8):
    init MLP, params; optimizer = Adam(lr=1e-3) + ReduceLROnPlateau
    repeat until early-stop on val log-data-loss:
        pred  = coverage(batch); 
        loss  = MSE(log10 pred, log10 label)         # data term (log space)
              + w_C * Cs_supervision_loss            # anchor C_s magnitude
              + w_m * monotonicity_penalty(dC_s/dU>0)
              + w_p * (log10 k_des - log10 k_des_prior)**2   # single-T only
        loss.backward(); optimizer.step()
# report mean +/- s.d. over seeds; identifiability via profile likelihood / Fisher
\end{verbatim}
\normalsize

\section*{Data Availability}
The data supporting the findings of this study were generated by simulation from
the Langmuir--Arrhenius ground truth and the reaction--transport model fully
specified in Section~4; all governing equations, boundary conditions, kinetic
parameters and sampling ranges are reported in the manuscript, so the datasets
(the single-temperature and multi-temperature CFD cases) can be regenerated
independently. The generated datasets and the associated PCINN training/inference,
identifiability-analysis (Fisher information, profile likelihood, initial-value
drift) and figure-generation code are available from the corresponding author upon
reasonable request.

\section*{Declaration of Generative AI and AI-Assisted Technologies in the Writing Process}
During the preparation of this work, the authors used AI-assisted tools to support
language editing, organization of manuscript drafts, and consistency checks across the
manuscript and supplementary files. The authors reviewed, edited, and verified all
AI-assisted output, including numerical claims and references, and take full
responsibility for the content of the publication.

\section*{Funding}
This work was supported by the State Key Laboratory of Ocean Engineering, Shanghai
Jiao Tong University (Grant No. GKZD010089), and the State Key Laboratory of
Acoustics, Chinese Academy of Sciences (Grant No. SKLA202406). The funders had no
role in the study design; collection, analysis, and interpretation of data; writing
of the manuscript; or decision to submit the article for publication.

\section*{Declaration of Competing Interest}
The authors declare that they have no known competing financial interests or personal
relationships that could have appeared to influence the work reported in this paper.

\section*{CRediT Authorship Contribution Statement}
Ning Hu: Conceptualization, Methodology, Software, Validation, Formal analysis,
Investigation, Data curation, Visualization, Writing -- original draft, Writing --
review and editing.

Chang Liu: Conceptualization, Methodology, Supervision, Resources, Funding
acquisition, Project administration, Writing -- review and editing.

Yunlei Jiang: Validation, Investigation, Writing -- review and editing.

Yuan Dong: Resources, Funding acquisition, Writing -- review and editing.

\section*{References}
\small
\sloppy
\begin{enumerate}[leftmargin=2.4em,labelindent=0pt,labelwidth=1.9em,labelsep=0.5em,align=left,itemsep=1pt,label={[\arabic*]}]
\item S.~M.~George, Atomic layer deposition: an overview, \textit{Chem. Rev.} 110 (2010) 111--131. \DOI{10.1021/cr900056b}.
\item P.~Poodt et al., Spatial atomic layer deposition: a route towards further industrialization, \textit{J. Vac. Sci. Technol. A} 30 (2012) 010802. \DOI{10.1116/1.3670745}.
\item R.~L.~Puurunen, Surface chemistry of atomic layer deposition: the trimethylaluminum/water process, \textit{J. Appl. Phys.} 97 (2005) 121301. \DOI{10.1063/1.1940727}.
\item I.~Langmuir, The adsorption of gases on plane surfaces of glass, mica and platinum, \textit{J. Am. Chem. Soc.} 40 (1918) 1361--1403. \DOI{10.1021/ja02242a004}.
\item A.~Yanguas-Gil, J.~W.~Elam, Analytic expressions for atomic layer deposition: coverage, throughput, and materials utilization, \textit{J. Vac. Sci. Technol. A} 32 (2014) 031504. \DOI{10.1116/1.4867441}.
\item M.~Raissi, P.~Perdikaris, G.~E.~Karniadakis, Physics-informed neural networks, \textit{J. Comput. Phys.} 378 (2019) 686--707. \DOI{10.1016/j.jcp.2018.10.045}.
\item G.~E.~Karniadakis et al., Physics-informed machine learning, \textit{Nat. Rev. Phys.} 3 (2021) 422--440. \DOI{10.1038/s42254-021-00314-5}.
\item A.~Raue et al., Structural and practical identifiability analysis of partially observed dynamical models by exploiting the profile likelihood, \textit{Bioinformatics} 25 (2009) 1923--1929. \DOI{10.1093/bioinformatics/btp358}.
\item C.~Rackauckas et al., Universal differential equations for scientific machine learning, arXiv:2001.04385 (2020).
\item M.~K.~Transtrum, B.~B.~Machta, J.~P.~Sethna, Geometry of nonlinear least squares with applications to sloppy models and optimization, \textit{Phys. Rev. E} 83 (2011) 036701. \DOI{10.1103/PhysRevE.83.036701}.
\item R.~T.~Q.~Chen, Y.~Rubanova, J.~Bettencourt, D.~Duvenaud, Neural ordinary differential equations, in: \textit{Advances in Neural Information Processing Systems (NeurIPS)}, 2018, pp.~6571--6583.
\item R.~N.~Gutenkunst, J.~J.~Waterfall, F.~P.~Casey, K.~S.~Brown, C.~R.~Myers, J.~P.~Sethna, Universally sloppy parameter sensitivities in systems biology models, \textit{PLoS Comput. Biol.} 3 (2007) e189. \DOI{10.1371/journal.pcbi.0030189}.
\item M.~K.~Transtrum, B.~B.~Machta, K.~S.~Brown, B.~C.~Daniels, C.~R.~Myers, J.~P.~Sethna, Perspective: sloppiness and emergent theories in physics, biology, and beyond, \textit{J. Chem. Phys.} 143 (2015) 010901. \DOI{10.1063/1.4923066}.
\item A.~Raue, M.~Schilling, J.~Bachmann, et al., Lessons learned from quantitative dynamical modeling in systems biology, \textit{PLoS ONE} 8 (2013) e74335. \DOI{10.1371/journal.pone.0074335}.
\item A.~F.~Villaverde, A.~Barreiro, A.~Papachristodoulou, Structural identifiability of dynamic systems biology models, \textit{PLoS Comput. Biol.} 12 (2016) e1005153. \DOI{10.1371/journal.pcbi.1005153}.
\item O.~Chi\c{s}, J.~R.~Banga, E.~Balsa-Canto, Structural identifiability of systems biology models: a critical comparison of methods, \textit{PLoS ONE} 6 (2011) e27755. \DOI{10.1371/journal.pone.0027755}.
\item M.~Temkin, V.~Pyzhev, Kinetics of ammonia synthesis on promoted iron catalysts, \textit{Acta Physicochim. URSS} 12 (1940) 327--356.
\item H.~Freundlich, \"{U}ber die Adsorption in L\"{o}sungen, \textit{Z. Phys. Chem.} 57 (1907) 385--470.
\item C.~E.~Rasmussen, C.~K.~I.~Williams, \textit{Gaussian Processes for Machine Learning}, MIT Press, Cambridge, MA, 2006.
\item M.~D.~McKay, R.~J.~Beckman, W.~J.~Conover, A comparison of three methods for selecting values of input variables in the analysis of output from a computer code, \textit{Technometrics} 21 (1979) 239--245. \DOI{10.1080/00401706.1979.10489755}.
\item B.~Efron, Bootstrap methods: another look at the jackknife, \textit{Ann. Statist.} 7 (1979) 1--26. \DOI{10.1214/aos/1176344552}.
\item D.~P.~Kingma, J.~Ba, Adam: a method for stochastic optimization, in: \textit{Int. Conf. on Learning Representations (ICLR)}, 2015. arXiv:1412.6980.
\item S.~Elfwing, E.~Uchibe, K.~Doya, Sigmoid-weighted linear units for neural network function approximation in reinforcement learning, \textit{Neural Networks} 107 (2018) 3--11. \DOI{10.1016/j.neunet.2017.12.012}.
\item L.~Lu, X.~Meng, Z.~Mao, G.~E.~Karniadakis, DeepXDE: a deep learning library for solving differential equations, \textit{SIAM Review} 63 (2021) 208--228. \DOI{10.1137/19M1274067}.
\item S.~L.~Brunton, J.~L.~Proctor, J.~N.~Kutz, Discovering governing equations from data by sparse identification of nonlinear dynamical systems, \textit{Proc. Natl. Acad. Sci. USA} 113 (2016) 3932--3937. \DOI{10.1073/pnas.1517384113}.
\item K.~Champion, B.~Lusch, J.~N.~Kutz, S.~L.~Brunton, Data-driven discovery of coordinates and governing equations, \textit{Proc. Natl. Acad. Sci. USA} 116 (2019) 22445--22451. \DOI{10.1073/pnas.1906995116}.
\item M.~Raissi, Deep hidden physics models: deep learning of nonlinear partial differential equations, \textit{J. Mach. Learn. Res.} 19 (2018) 1--24.
\item O.~Owoyele, P.~Pal, ChemNODE: a neural ordinary differential equations framework for efficient chemical kinetic solvers, \textit{Energy and AI} 7 (2022) 100118. \DOI{10.1016/j.egyai.2021.100118}.
\item C.~Kreutz, A.~Raue, D.~Kaschek, J.~Timmer, Profile likelihood in systems biology, \textit{FEBS J.} 280 (2013) 2564--2571. \DOI{10.1111/febs.12276}.
\item G.~Pang, L.~Lu, G.~E.~Karniadakis, fPINNs: fractional physics-informed neural networks, \textit{SIAM J. Sci. Comput.} 41 (2019) A2603--A2626. \DOI{10.1137/18M1229845}.
\item S.~Cai, Z.~Mao, Z.~Wang, M.~Yin, G.~E.~Karniadakis, Physics-informed neural networks (PINNs) for fluid mechanics: a review, \textit{Acta Mech. Sin.} 37 (2021) 1727--1738. \DOI{10.1007/s10409-021-01148-1}.
\item R.~G.~Patel, I.~Manickam, N.~A.~Trask, et al., Thermodynamically consistent physics-informed neural networks for hyperbolic systems, \textit{J. Comput. Phys.} 449 (2022) 110754. \DOI{10.1016/j.jcp.2021.110754}.
\item E.~Kharazmi, Z.~Zhang, G.~E.~Karniadakis, hp-VPINNs: variational physics-informed neural networks with domain decomposition, \textit{Comput. Methods Appl. Mech. Engrg.} 374 (2021) 113547. \DOI{10.1016/j.cma.2020.113547}.
\item M.~Raissi, A.~Yazdani, G.~E.~Karniadakis, Hidden fluid mechanics: learning velocity and pressure fields from flow visualizations, \textit{Science} 367 (2020) 1026--1030. \DOI{10.1126/science.aaw4741}.
\item A.~Yazdani, L.~Lu, M.~Raissi, G.~E.~Karniadakis, Systems biology informed deep learning for inferring parameters and hidden dynamics, \textit{PLoS Comput. Biol.} 16 (2020) e1007575. \DOI{10.1371/journal.pcbi.1007575}.
\item S.~Wang, X.~Yu, P.~Perdikaris, When and why PINNs fail to train: a neural tangent kernel perspective, \textit{J. Comput. Phys.} 449 (2022) 110768. \DOI{10.1016/j.jcp.2021.110768}.
\item A.~D.~Jagtap, K.~Kawaguchi, G.~E.~Karniadakis, Adaptive activation functions accelerate convergence in deep and physics-informed neural networks, \textit{J. Comput. Phys.} 404 (2020) 109136. \DOI{10.1016/j.jcp.2019.109136}.
\item F.-G.~Wieland, A.~L.~Hauber, M.~Rosenblatt, C.~T\"{o}nsing, J.~Timmer, On structural and practical identifiability, \textit{Curr. Opin. Syst. Biol.} 25 (2021) 60--69. \DOI{10.1016/j.coisb.2021.03.005}.
\item D.~Pan, L.~Ma, Y.~Xie, T.~C.~Jen, C.~Yuan, On the physical and chemical details of alumina atomic layer deposition: a combined experimental and numerical approach, \textit{J. Vac. Sci. Technol. A} 33 (2015) 021511. \DOI{10.1116/1.4905726}.
\item M.~K.~Transtrum, P.~Qiu, Model reduction by manifold boundaries, \textit{Phys. Rev. Lett.} 113 (2014) 098701. \DOI{10.1103/PhysRevLett.113.098701}.
\item C.~Masse de la Huerta, V.~H.~Nguyen, J.-M.~Dedulle, D.~Bellet, C.~Jim\'enez, D.~Mu\~noz-Rojas, Influence of the geometric parameters on the deposition mode in spatial atomic layer deposition: a novel approach to area-selective deposition, \textit{Coatings} 9 (2019) 5. \DOI{10.3390/coatings9010005}.
\item Z.~Deng, W.~He, C.~Duan, R.~Chen, B.~Shan, Mechanistic modeling study on process optimization and precursor utilization with atmospheric spatial atomic layer deposition, \textit{J. Vac. Sci. Technol. A} 34 (2016) 01A108. \DOI{10.1116/1.4932564}.
\item D.~Pan, T.~C.~Jen, C.~Yuan, Effects of gap size, temperature and pumping pressure on the fluid dynamics and chemical kinetics of in-line spatial atomic layer deposition of Al$_2$O$_3$, \textit{Int. J. Heat Mass Transf.} 96 (2016) 189--198. \DOI{10.1016/j.ijheatmasstransfer.2016.01.034}.
\end{enumerate}

\end{document}